\documentclass[authoryear,preprint,review,11pt]{elsarticle}
\usepackage[left=1in,right=1in,top=1in,bottom=1in]{geometry}
\usepackage{multirow, booktabs}
\usepackage{graphicx, caption, subcaption}
\usepackage{tabularx}

\usepackage{soul} 
\usepackage{xcolor}
\sethlcolor{yellow}

\usepackage{amsmath}
\usepackage{mathrsfs}
\usepackage{optidef}
\usepackage{ifthen}

\usepackage[section]{placeins}
\usepackage{verbatim}
\newlength{\imagewidth}
\newlength{\multilen}
\newboolean{twocolumnmode}
\setboolean{twocolumnmode}{false} 

\usepackage{amssymb}
\usepackage{lmodern}
\usepackage{comment}
\usepackage{hyperref}
\hypersetup{
    colorlinks=true,
    linkcolor=blue,
    filecolor=magenta,      
    urlcolor=blue,
}
 
\journal{arXiv}

\begin{document}

\begin{frontmatter}


 \title{A Distributionally Robust Optimisation Approach to Fair Credit Scoring\tnoteref{t1}}
 \author[1]{Pablo Casas
 }
  \ead{pablo.casas@soton.ac.uk}
 \author[1]{Huan Yu\corref{cor1}}
 \ead{huan.yu@soton.ac.uk}
  \author[1]{Christophe Mues
 }\ead{c.mues@soton.ac.uk}

 \cortext[cor1]{Corresponding author.}
\affiliation[1]{
            addressline={Southampton Business School}, 
            organization={University of Southampton},
            city={Southampton},
            postcode={SO16 7QF}, 
            country={United Kingdom}}




\begin{abstract}
Credit scoring has been catalogued by the European Commission and the Executive Office of the US President as a high-risk classification task, in light of the potential 
harms of making loan approval decisions based on models that would be biased against certain groups. To address this concern, recent credit scoring research has considered a range of fairness-enhancing techniques put forward by the machine learning community to reduce bias and unfair treatment in classification systems.  
While the definition of fairness or the approach they follow to impose it may vary, most of these techniques, however, disregard the robustness of the results. This can create situations where unfair treatment is effectively corrected in the training set, but when producing out-of-distribution classifications, unfair treatment is incurred again.
Instead, in this paper, we will investigate how to apply Distributionally Robust Optimisation (DRO) methods to credit scoring, thereby empirically evaluating how they perform in terms of fairness, ability to classify correctly, and the robustness of the solution against changes in the marginal proportions.
In so doing, we find DRO methods to provide a substantial improvement in terms of fairness, with almost no loss in predictive performance. These results thus indicate that DRO can improve fairness in credit scoring, provided that further advances are made in efficiently implementing these systems. 
In addition, our analysis suggests that many of the commonly used fairness metrics are not ideally suited to the credit scoring setting, as they evaluate performance at a single classification threshold. 
\end{abstract}

\end{frontmatter}

\section{Introduction}
\label{intro}
As Machine Learning (ML) is becoming increasingly common in the financial sector \citep{garcia2015creditf}, there are rising concerns regarding its use and fairness of treatment in settings such as credit scoring, where ML may be used to screen individual loan applications by predicting their risk of default.
Credit scoring has been catalogued as a high-risk automated decision by governing bodies such as the European Commission or the Executive Office of the US President \citep{falque-pierrotin_2017, muñoz_smith_patil_2016}. This reflects concerns that any unfair treatment in this context could be especially damaging to the public, potentially exacerbating wealth discrepancy between groups. Hence, lenders are expected to ensure that the credit scoring systems they build do not produce loan decisions that exhibit undesirable discriminatory bias, placing certain protected (demographic or ethnic) groups at a systematic disadvantage in terms of 
access to credit or the costs involved. This makes work on ML fairness that focuses 
on credit scoring applications relevant for both regulatory and ethical reasons.

Unfair treatment by ML systems is often due to biases in the data. More specifically, the sources of bias in systems are often divided into four \citep{pessach2020algorithmic}: missing data, objective function, training data, and proxy attributes. Firstly, bias can be caused by missing data if, for example, a protected group has more missing values than the non-protected group, allowing the algorithm to make biased decisions stemming from treating those values \citep{martinez2019fairness}. Secondly, having been trained to minimise an error metric or loss function over the entire training sample, the algorithm might underperform on minority groups, as their smaller size implies that errors made here have limited impact on average performance \citep{pessach2020algorithmic}. This can be exacerbated by the severe class imbalance that is often present in credit scoring settings, which could make it harder to effectively distinguish between high-risk and low-risk protected applicants. Lastly, attributes that correlate with sensitive attributes may act as proxies for those attributes, and become an indirect source of discrimination. For example, attributes such as income, employment, and marital status can be correlated to gender \citep{selbst2017disparate}, or combining multiple features, such as address with name or other features, could even reveal information about an individual's ethnicity \citep{chouldechova2018frontiers}.

Although there is some agreement among academics and practitioners on these sources of bias, how to define fairness remains a subject of ongoing debate. Different scholars have proposed different ways of categorising unfairness metrics; in this paper, we will use the independence, separation, and sufficiency distinction used by \citet{barocas2017fairness} and employ measures related to separation.

Intuitively, leaving the sensitive attribute (and, possibly, its close proxies) out of the model might be seen as one way of trying to tackle the problem. Indeed, this has traditionally been the standard practice and enforced by law \citep{andreeva2019law}. However, when doing so, one can fall into the trap of omitted variable bias. \citet{andreeva2019law} exemplified how using a model where gender had been excluded as a variable further increased the prejudicial bias towards female applicants in a credit scoring context, hence suggesting that sensitive attributes may need to be part of the scorecard/model development process, at least in its training phase. 
 
In credit scoring, several methods from the fairness-aware ML literature have been applied to reduce unfair treatment, such as reweighting \citep{calders2009building}, the prejudice remover \citep{kamishima2011fairness}, and reject option classification \citep{kamiran2012decision}. 
To empirically test their effectiveness, this work often selects the applicant's age as the sensitive attribute, since gender or ethnic data are rarely included in publicly available credit scoring datasets
. Younger applicants are then usually designated as the protected group \citep{kamiran2009classifying}.

A limitation in many of these studies, however, is that they do not consider out-of-distribution performance on any more recent data collected after model development. The credit scoring case is especially interesting in that regard, as the data we use to train the model can significantly differ from the current data, particularly as the true label (default/not default) is not revealed immediately but rather after a long delay. Indeed, out-of-sample fairness can worsen when training a fair classifier without incorporating an out-of-sample guarantee \citep{wang2020robust}.
Any such changes in the data distribution can be caused by a variety of factors: noise in the data, as simulated by \citet{wang2020robust}; major economic or demographic events, such as a crisis or a pandemic, which may cause sudden population shifts \citep{ovadia2019can}; more progressive changes of the data over time, which are generally referred to as population drift \citep{krempl2011classification}; or changes in the relations between the data and the outcome, which is referred to as concept drift \citep{vzliobaite2016overview}. Alternatively, these shifts may stem from operational changes, such as updated screening procedures or internal credit policies, which could significantly alter the evaluation population.

One solution to the out-of-distribution performance issue is to use Distributionally Robust Optimisation (DRO), which can limit the decrease in accuracy in the presence of data shifts \citep{subbaswamy2021evaluating}. However, most DRO research focuses exclusively on robustness for predictive performance, whilst leaving fairness out of scope. 
The argument for including fairness is that this could make the model more robust to distributional uncertainty not only in the target but also, when explicitly modelled, in the sensitive attributes, as explored by \citet{wang2020robust}. A related idea has been incorporated in the 
classification system developed by \citet{taskesen2020distributionally}, 
which adds a regularisation term that penalises unfair treatment (measured by a convex approximation of the equalised odds criterion) to their distributionally robust Logistic Regression model. 

Hence, in this paper, we posit that DRO with fairness constraints may provide a suitable solution 
to enhance both the robustness and fairness of credit scoring, the effectiveness of which we then test on public data. 
Such a solution requires that the modeller decides on the trade-off between optimality and robustness, which has been one of the main foci in robust optimisation \citep{bertsimas2004price, ben1998robust}.
To assist with this, our paper contains a section specifying the different hyperparameters and their respective trade-offs.

This paper will, therefore, address the following main questions. First, is there any added benefit of using DRO in a credit scoring setting? Second, what are the practical modelling aspects that a credit scorer should consider when using DRO? 

To answer these, we will first review the related literature and then present the different methods included in our study: classic Logistic Regression (LR)---with and without regularisation; distributionally robust LR (DRLR); fair LR (FLR); and distributionally robust fair LR (DRFLR). Next, we will describe the datasets used and how we have treated them, and then compare and analyse the performance of the different models trained on them. The paper will conclude by discussing some key insights gained and suggesting ideas for further research. 

\section{Related literature}
\label{lit}
Dealing with data or distributional uncertainty presents challenges for modellers in many application settings, as they may lack precise data on certain problem parameters or knowledge of their true distributions. 
In such cases, they may favour solutions that perform well under a variety of possible scenarios (i.e.\@ realisations of those uncertain problem parameters) or possible probability distributions. Similarly, seeking a degree of robustness (as opposed to simply maximising utility
) appears a common trait of human decision making under uncertainty more generally \citep{long2023robust}. Two well-known classes of optimisation methods that have been developed to tackle different types of uncertainty are Stochastic Optimisation (SO) and (classical) Robust Optimisation (RO).\footnote{Another method that has been proposed is chance-constrained optimisation. This assumes an underlying distribution and constrains the optimisation system based on confidence intervals. While this approach is also popular, it is beyond the scope of this paper.}
 
To handle probabilistic uncertainty, SO methods treat some problem parameters as random variables with known or approximate probability distributions. Where the true parameters of these distributions are unknown, SO methods assume that past or current data 
are representative of future conditions and can, therefore, be used to estimate them
. Hence, they seek to optimise the expected value of the objective function taken over those distributions. 
A related principle underlies most supervised ML systems
, which can be understood as aiming to minimise the expected value of a loss function penalising inaccurate predictions. 
This expected loss is then commonly approximated by the empirical loss computed from the training data; if the latter is not a good representation of the true distribution, the resulting model may underperform on future unseen data. 

Rather than employing probabilistic information, RO, introduced by \citet{soyster1973convex}, 
looks to robustify the solution of an optimisation problem against uncertainty in the data by allowing the modeller to specify a range of possible values (referred to as the \textit{uncertainty set}). As argued by \citet{bertsimas2004price}, though, as far as the trade-off between performance and robustness is concerned, RO tended to give up too much performance, as its idea was to consider all possible realisations in this set and find a solution that performs well for the worst-case scenario. 

Instead of assuming a specified distribution for the uncertain problem parameters (as in SO) or an uncertainty set over their values (as in RO), DRO models uncertainty about the true underlying distribution by requiring it to lie ``near'' the empirical (or other nominal) distribution. 
DRO then optimises over the worst possible distribution within this so-called \textit{ambiguity set} of neighbouring distributions. 
%
%
%
%
This idea appears promising for credit scoring applications as it might limit performance loss in the event of (minor) distributional shifts. Furthermore, DRO can be readily applied to LR, which, regardless of the (often rather small) performance differential relative to more recent ML algorithms, is still a widely applied method in credit scoring practice \citep{lessmann2015benchmarking}. This enduring popularity is likely linked to the interpretability advantages that LR offers over more complex systems in such a highly regulated setting. Therefore, we have chosen to focus on (generalised) linear models in this paper. 
 

Interestingly, there is a direct connection between DRO and the concept of regularisation \citep{rahimian2019distributionally}. In ML, the latter is commonly applied to prevent overfitting and improve the algorithm's ability to generalise beyond the training data \citep{tian2022comprehensive, salehi2019impact}. More specifically, it has been shown that the two most common regularisation methods, ridge and lasso, admit distributionally robust interpretations that connect them to robustness against uncertainty exhibiting Gaussian-like or Laplacian-like structure, respectively \citep{blanchet2019robust}. These can thus be seen as special cases, whilst DRO allows one to consider a broader range of distributions that are sufficiently close to the empirical distribution \citep{shafieezadeh2020wasserstein}. It has also been argued that (heuristic) regularisation methods are less reliable than the robustness implied by DRO \citep{gao2022wasserstein, xu2010robust}.\footnote{It is worth mentioning here that DRO is often also interpreted through game theory or risk aversion \citep{rahimian2019distributionally} rather than viewed as a form of regularisation.}
 
To construct ambiguity sets, one needs some method to determine how close or similar a given distribution is to the empirical distribution. These methods are generally divided into two categories: moment-based (those that consider similar distributions) and distance/metric-based (which consider close distributions). Note, however, that there are other ways of dividing the types of ambiguity sets \citep{rahimian2019distributionally}. Amongst the distance-based metrics, the choice will depend on the task at hand, as different metrics have different effects \citep{cuturi2014ground}. In this paper, we have opted to use Wasserstein distance for several reasons: it has a clear intuition, making it more approachable for practitioners; it has been shown to have stochastic approximations that could be computed efficiently; and it allows for customisation of the concept of distance.

This paper does not focus solely on robustness, but also examines how DRO could contribute to enhancing fairness. 
The idea behind including an unfairness penalty within a DRO system is to create systems that perform better in the presence of uncertainty in the sensitive attribute. This has been previously explored by \citet{taskesen2020distributionally}, as well as by \citet{wang2020robust}. The latter empirically exemplified (on a selection of 
small benchmarking datasets from a variety of application domains) that treatment tends to be more unfair when a non-robust approach is used. Hence, merely adding an unfairness penalty term to non-robust methods 
might not be sufficient. However, these methods have not yet been trialled more extensively in the credit scoring setting. 

Interestingly, 
\citet{slowik2021algorithmic} provide a fundamental critique of this algorithmic focus; they mathematically demonstrate that, for practical purposes, DRO is equivalent to a data-reweighting algorithm. This implies that instead of relying on complex algorithmic penalties, fairness goals might be addressed more directly through deliberate data curation. 

To be able to detect, let alone treat unfairness, one has to decide how to measure it. Although there are various ways of categorising the multitude of fairness measures, one of the most popular classifications, proposed by \citet{barocas2017fairness}, distinguishes between three groups: \emph{independence} measures, that use the predicted outcome and the sensitive attribute (e.g.\@ statistical parity) \citep{dwork2012fairness}; \emph{separation} measures, which require that certain performance metrics are equal across protected and non-protected groups (e.g.\@ equalised odds) \citep{NIPS2016_9d268236}; and \emph{sufficiency} measures, that use the probability of the outcome (e.g.\@ matching conditional frequencies) \citep{chouldechova2017fair}. Methods to enhance these notions of fairness may then be applied to the pre- \citep{zhang_review_2023}, in- \citep{wan_-processing_2023} or post-processing \citep{petersen_post-processing_2021} stages of model development. Those may involve the use of a procedure to de-bias the data, incorporating one of the aforementioned fairness criteria during model training itself, or adjusting how the model outputs are employed to make decisions, respectively.

One extensive study of fairness in credit scoring has been undertaken by \citet{kozodoi2022fairness}, from which two key takeaways are that separation measures appear to be the ones best suited for credit scoring, and that in-processing techniques tend to yield the largest improvements in fairness whilst having limited impact on profit (making them the most efficient in the authors' Pareto analysis). In addition, these techniques offer the flexibility of being able to decide on the trade-off between performance and fairness through hyperparameter tuning. This capacity to systematically navigate the Pareto frontier closely complements the framework introduced by \citet{hurlin2026fairness}, who utilise feature-level significance testing to directly optimise the fairness-performance trade-off for risk-screening algorithms. Indeed, the topic of fairness in credit scoring has attracted substantial interest in recent years, with numerous studies expanding on these optimisation challenges \citep{kim2026fair,sargeant2025formalising,11401835}.


In summary, the literature above suggests 
a need for fair and robust methods in credit scoring and 
further investigation of the effects of parameter tuning on the fairness of the scoring. In the next section, after introducing our notation and traditional logistic regression and its regularised version, 
we will explain distributionally robust logistic regression, with or without fairness constraints. 

\section{Methods}
\label{methods}
Credit scoring aims to predict whether a potential borrower is likely to default, thus giving an estimate of the level of risk to the lender. To build their credit scoring model, the lender takes data from previous borrowers. 
Each such past application $i$ is represented by a feature vector $\mathbf{x}_i \in \mathbb{R}^m$, an observation of the input variable $X$ containing applicant or loan-level data available at that time. %
Let there be $n$ observations in the sample set
, the default outcomes for all of which we denote by \(\mathbf{y}\). 
For each $\mathbf{x}_i$, we thus observe the true label $y_i$, the realisation of a binary variable $Y$ where 1 represents a bad payer (default), and 0 a good payer (not defaulted). 
We introduce another variable $S$ to denote the binary sensitive attribute, which is not part of the $m$-dimensional feature space. 
By $\mathbf{s}$, we denote a vector of size $n$ containing values of 1 for the observations that belong to the protected group, and 0 for those in the non-protected group. For example, in the case of potential age discrimination towards young applicants, ones will be assigned to those younger than the chosen age threshold, while zeros will represent those above it. 
Let us, for now, disregard the information in $\mathbf{s}$ and start by developing a group-agnostic model. 

A model now needs to be trained to produce a series of predictions, \(\hat{\mathbf{y}}\), for $\mathbf{y}$, which could then be used by the lender to estimate the Probability of Default (PD) of any new applicants. Therefore, the traditional credit scoring problem can be formulated as follows: \\ Given the matrix of observations (inputs), \vspace{+3pt}\\
\vspace{+4pt} \(
\mathbf{X} = \begin{pmatrix} 
    \mathbf{x}_{1}^\top \\
    \vdots \\
    \mathbf{x}_{n}^\top  
    \end{pmatrix} = 
\begin{pmatrix} 
    x_{11} & \dots  & x_{1m}\\
    \vdots & \ddots & \vdots\\
    x_{n1} & \dots  & x_{nm} 
    \end{pmatrix}
\), and the default information (target),  \( \mathbf{y} = \begin{pmatrix} 
    y_{1} \\
    \vdots \\
    y_{n} 
    \end{pmatrix}   
\),
we seek a function $h(\cdot)$ mapping inputs to predictions, $\hat{y}_i = h(\mathbf{x}_i)$ which closely match the true outcomes $y_i$, that is, one that minimises some chosen loss function, $\ell(y_i,\hat{y}_i)$, over the set of training examples, $i=1,...,n$. 
Since, for the reasons given in section \ref{lit}, we restrict ourselves to (generalised) linear prediction models in this paper, this requires finding an optimal vector of coefficients or weights, \(\mathbf{w} = \bigl( w_{1} \, \dots \, w_{m}\bigr)^{\!\top}\!\), for a function $h(\mathbf{x}; \mathbf{w})$ that applies some transform $f(\cdot)$ to the weighted sum of inputs; i.e.\@ we let \( h(\mathbf{x}_i;\mathbf{w}) = f(\mathbf{w}^{\!\top}\mathbf{x}_i) \). For example, if we were to pick the identity function and let $f(v)\mathbin{=}v$, the output would be modelled as a simple linear function of the inputs. 

To solve this problem, mathematical programming techniques such as linear programming have been proposed \citep{thomas_edelman_croock_2002}, but a more widely used method currently is to build a binary classification model using logistic regression. This uses the logistic transform, i.e.\@ $f(v) = \frac{1} {1 + e^{-v}}$, to rescale the weighted sum of inputs to the $(0,1)$ range.  
Hence, the logistic regression model assumes the following form: 
\begin{equation}
  \label{scoringfunc}  
 h(\mathbf{x};\mathbf{w}) = (1+e^{-\mathbf{w}^\top \mathbf{x}})^{-1}.
 \end{equation}
For each observation $i$, we consider the \textit{log loss}; i.e.\@ we let 
\begin{equation}
\label{logloss}
\ell(y_i,\hat{y}_i) = - \big(y_i \ln(\hat{y}_i) + (1-y_i)\ln(1-\hat{y}_i)\big), \text{ with } \hat{y}_i = h(\mathbf{x}_i;\mathbf{w}).
\end{equation}
\noindent The classic training objective then is to find the weights that minimises the mean log loss over the sample of observations. This model has proven attractive to credit scorers as it is easy to interpret and has produced consistently good predictive performance (especially among other non-ensemble models) in many empirical studies \citep{moscato2021benchmark, lessmann2015benchmarking}. Furthermore, it offers a probabilistic interpretation, as $\hat{y}_i$ is the model's estimate of the probability that applicant $i$ is a bad payer. This enables the lender to make a more informed decision on where to set their application score threshold. 
For example, risk-averse lenders might only accept applications whose default risk is very low, whereas those with a higher risk appetite might opt for a higher cut-off. 
 
In summary, the optimisation problem that logistic regression is aiming to solve is as follows. \\
For a training sample of observations \((\mathbf{x}_i,y_i),\, i=1\dots n\), we seek a weight vector $\mathbf{w}$ that minimises the empirical expected log loss
; that is,  
\begin{equation} \label{LR_emp_loss_min}
\underset{\mathbf{w} \in \mathbb{R}^m}{\min} \; \frac{1}{n} \underset{i=1}{\overset{n}{\sum}}\ell\big(y_i,h(\mathbf{x}_i; \mathbf{w})\big). 
\end{equation}
Once the solution is found, we can apply the resulting 
scoring function (\ref{scoringfunc}) to a new application, $\mathbf{x}$, with $m$ feature values but no information yet on the outcome $y$, to obtain 
its PD, i.e.\@ \(\hat{y}=h(\mathbf{x};\mathbf{w})\). 

One of the common issues with \eqref{LR_emp_loss_min} is that it assumes that the empirical distribution of the training data 
(denoted by $\hat{\mathbb{P}}_n$) is a good approximation of the underlying true distribution $\mathbb{P}$ of $(X,Y)$. Where this does not hold, such as when dealing with smaller sample sizes, or when experiencing population drift or changes in economic conditions, this can easily result in overfitting or poor performance. One solution to the overfitting problem may lie in including a regularisation constraint, \(\Omega(\mathbf{w})\), to penalise the model for becoming too complex and thus reduce overfitting. More formally, such approach looks to solve:
\begin{mini}|s|
{\scriptstyle \mathbf{w} \in \mathbb{R}^m}{ \frac{1}{n} \underset{i=1}{\overset{n}{\sum}}\ell\big(y_i,h(\mathbf{x}_i;\mathbf{w})\big) }
{\label{LR_loss_min_w_regul}}{}
\addConstraint{\Omega(\mathbf{w}) \leq C },
\end{mini}
where $C$ is a constant that indicates the relative weight given to the regularisation component. The Lagrangian of this problem leads to the following regularised logistic regression formulation: 
\begin{equation} \label{regul_log_reg}
\underset{\mathbf{w}\in \mathbb{R}^m}{\min}  \; \; \frac{1}{n} \underset{i=1}{\overset{n}{\sum}}\ell\big(y_i,h(\mathbf{x}_i;\mathbf{w})\big) + \lambda \Omega(\mathbf{w}). 
\end{equation}

\noindent Although there are various types of regularisation \citep{tian2022comprehensive}, two common forms are to add either an 
L1 (lasso) or L2 (ridge) penalty term, which uses the 1-norm or squared 2-norm of the weight vector, respectively. For example, to implement L2 regularisation, we insert the following penalty: 
\begin{equation} \label{ridge_pen}
\begin{aligned}
\Omega(\mathbf{w})= ||\mathbf{w}||_2=\overset{m}{ \underset{i=1}{\sum}}\;w_i^2. 
\end{aligned}
\end{equation} 
These two regularisation methods can be understood as assuming a Laplacian or Gaussian prior, respectively, on the regression coefficients \citep{murphy2012machine}. In ML practice, however, without such an explicit Bayesian justification, and as they require further empirical tuning, they are often employed as heuristic methods for improving generalisation \citep{ying2019overview}. 

Whereas the above methods can help control model complexity, they do not account for data uncertainty. In a classification setting such as ours, however, there may be uncertainty over the training observations. 
For example, an applicant’s true income value may deviate from that which is reported. To make solutions robust in the presence of such uncertainty, classical RO considers the worst-case scenario 
over an uncertainty set, 
\( \mathcal{U}= \{ \mathbf{X}\!+\!\Delta \mid \Delta\!\in\!\mathcal{D}\!\subseteq\!\mathbb{R}^{n \times m} \} \), where $\mathcal{D}$ is a set of admissible perturbations to the training data.\footnote{We consider here robustness against uncertainty in features, but one could similarly introduce the notion of robustifying classifiers against uncertainty in the output labels of the training set.} 
This set is pre-specified by the modeller through the choice of a structural form, which could range from a box uncertainty set (which simply assumes that each feature can vary independently within fixed bounds) to more sophisticated alternatives. With the uncertainty set defined, feature weights can then be sought by solving 
the corresponding optimisation problem for any of the data in this range. Thus, RO aims to ensure that the solution does not perform poorly for any data that differs from the observed training data in the manner permitted by the chosen uncertainty set. 
%
%

A general formulation of the RO counterpart to binary classification is developed in Chapter 12 of \citet{ben2009robust}. Using the log loss function, this can be reformulated as: 
%
%
%
%
\begin{equation} \label{RO_classif}
\underset{\mathbf{w} \in \mathbb{R}^m}{\min}\;
\underset{(\tilde{\mathbf{x}}_1\,\dots\,\tilde{\mathbf{x}}_n)^{\!\top} \! \in \, \mathcal{U}}{\max}\;\;
\frac{1}{n} \sum_{i=1}^{n} \ell\big(y_i,\,h(\tilde{\mathbf{x}}_i;\mathbf{w})\big).
\end{equation}
%
%
\noindent Hence, one would seek weights that minimise the worst-case loss over the uncertainty set. 
The over-pessimistic nature of this framework commonly results in underperformance. Nonetheless, it could be useful in settings where data is not exactly known and large errors should be avoided at all costs. 
This is not the general setting in credit scoring, however, as a lender would rather favour a model that performs consistently well, both on the data at hand, as well as in the presence of uncertainty or variations in the underlying population distribution. 

To help balance those criteria, DRO seeks to minimise the worst-case expected loss over a set of distributions that are close to a nominal distribution---typically the empirical distribution. Hence, rather than simply minimising the empirical loss (like in standard training) or defining an uncertainty set that disregards distributional information (as classical RO does), DRO accounts for distributional uncertainty by introducing a so-called ambiguity set of nearby distributions.\footnote{Note again the distinction between ambiguity and uncertainty set; the former is a set of distributions while the latter is a set of scenarios/possible observations.} 
%
This feature makes DRO attractive not just to credit scorers, as \citet{blanchet2019robust} shows that DRO can have a variety of ML applications. 


DRO can be made to emulate some of the characteristics of either classical RO or SO, making it a more flexible alternative. 
Specifically, by increasing the parameter \(\rho\) (which controls the ambiguity set's size), DRO becomes more conservative and can be made to approach a worst-case formulation similar to classical RO. 
Conversely, as \(\rho\) is shrunk towards zero, the ambiguity set collapses to a single distribution, reducing the problem to the standard sample average approximation of SO. 
With values for \(\rho\) set somewhere in between these extremes, we can instead account for a moderate level of conservatism.

As noted in section \ref{lit}, to construct a DRO problem,  one needs to define the ambiguity set, i.e.\@ the set of probability distributions that we wish to consider as we look for the worst possible distribution. In this paper, we employ a distance-based metric, specifically the Wasserstein metric \citep[originally introduced by][]{kantorovich1960mathematical}, to create the ambiguity set. This approach offers several advantages to credit scoring practitioners. 
It allows them to vary the contribution of different variables to the distance calculations, depending on which distributional deviations they consider more plausible. For instance, if larger shifts are anticipated in the distribution of default outcomes than in the application features, this could, in principle, be accounted for in the ground metric used. 
Wasserstein ambiguity sets also have a direct connection to regularisation \citep{gao2022wasserstein}, which can make the method more intuitive to credit scorers. Furthermore, for logistic regression, using the Wasserstein metric in combination with the log loss function yields a convex optimisation problem, which is computationally more tractable and can be solved with an exponential cone solver or approximated as a linear programming problem \citep{li2019first}.
 
The type-1 Wasserstein distance (also commonly named Earth Mover's distance) between two distributions \( \mathbb{Q}_1\) and \( \mathbb{Q}_2\) is defined as:
%
%
 \begin{equation} \label{wasserstein_distance}
 \mathrm{W}(\mathbb{Q}_1, \mathbb{Q}_2) = \inf_{\pi \in \Pi(\mathbb{Q}_1, \mathbb{Q}_2)} \mathbb{E}_{(\xi, \xi') \sim \pi} \big[c(\xi, \xi')\big].
\end{equation}
%
\noindent
Here, \(\Pi(\mathbb{Q}_1,\mathbb{Q}_2)\) denotes the set of all joint distributions whose marginals are \(\mathbb{Q}_1\) and \(\mathbb{Q}_2\),  
and $c(\cdot\,,\cdot)$ is the ground metric, i.e.\@ our measure of distance between any pair of coupled points; for example, choosing the 2-norm yields Euclidean distance. 
Note that whereas the ground metric measures distance between points, the Wasserstein distance is the minimal expected cost of ``transporting" the observations of one distribution to another, given that ground metric
. It is worth mentioning that this measure is accompanied by an optimal transport plan that may reveal the regions where two distributions differ most.

To make our distributionally robust approach fairness-aware, we incorporate both the sensitive attribute $S$ and the outcome $Y$ into the observation space over which the Wasserstein distance \eqref{wasserstein_distance} is measured. 
Each point is thus represented as $(\mathbf{x},s,y)$, with the ground metric defined over this extended space, and $\hat{\mathbb{P}}_n$ now redefined to denote its empirical distribution. Specifically, 
we will employ {\citeauthor{taskesen2020distributionally}}'s (\citeyear{taskesen2020distributionally}) ground metric, also used by \citet{wang2024wasserstein}, which measures the distance between such points as the sum of the norm of the difference between their feature vectors and a weighted sum of the absolute difference between their sensitive and target attributes, 
\begin{equation}\label{ground_metric}
c\big((\mathbf{x},s,y),(\mathbf{x}',s',y')\big)= ||\mathbf{x}-\mathbf{x}'||+ \kappa_s|s-s'|+\kappa_y|y-y'|.
\end{equation} 
\noindent
This choice of ground metric allows for different transportation costs\footnote{As the Wasserstein distance comes from the field of optimal transport, the variables that control the relative importance of each feature when measuring distances are named transportation costs.\vspace{+6pt}}, \(\kappa_s\) and \(\kappa_y\), for the sensitive attribute and the assigned class, respectively. These parameters can be intuitively interpreted as regulating the relative trust placed in $s$ and $y$, respectively. In this context, this notion of relative trust serves as a conceptual proxy for the modeller's confidence in said observations. 

 
From \eqref{ground_metric}, we can define an ambiguity set \(\mathcal{B}_\rho(\mathbb{\hat{P}}_n)\) consisting of all probability distributions that satisfy a maximum discrepancy condition with respect to the nominal (empirical) distribution, 
\begin{equation}\label{wasserstein_ball}
\mathcal{B}_\rho(\mathbb{\hat{P}}_n)= \big\{\mathbb{Q}\in \mathcal{F}:\mathrm{W}(\mathbb{Q},\mathbb{\hat{P}}_n) \leq \rho \big\},
\end{equation} 
where \(\mathcal{F}\) is the set of all possible distributions over $(X, S, Y)$
, and $\rho$ specifies the maximum Wasserstein distance allowed. Hence, $\rho$ can be seen as representing the level of distrust in the data, \(\mathbb{\hat{P}}_n\), while \( \kappa_s\) and \(\kappa_y\) are a way of parameterising how much $S$ and $Y$ each contribute to that (dis)trust. 
The set defined by Equation \eqref{wasserstein_ball} is often called a Wasserstein ball, as it contains all distributions within a Wasserstein distance \(\rho\) (the radius) from the nominal distribution (the centre of the ball). 
This allows us to put forward the following DRO problem formulation: 
%
%
\begin{equation} \label{DRO_classifier}
\underset{\mathbf{w} \in \mathbb{R}^m}{\min} \;
\underset{\mathbb{Q} \in \mathcal{B}_\rho(\hat{\mathbb{P}}_n)}{\sup} \;
\mathbb{E}_{
\mathbb{Q}} \big[ 
    \ell\big(Y, h(X;\mathbf{w})\big) 
\big].
\end{equation}


This nested optimisation problem simultaneously considers two objectives: one is to look for the distribution $\mathbb{Q}$ within the Wasserstein ball that maximises the expected loss; the other is to find the weights that minimise that loss. 
In our setting, one can think of its solution as the model whose performance holds up best under the worst deviations from the training data distribution permitted by the ambiguity set.\footnote{This can be understood as a game whereby one player tries to maximise performance whilst the other tries to hinder it; this way, the optimisation is forced to produce a robust solution for which perturbations are not detrimental; hence, DRO is sometimes linked to game theory \citep{rahimian2019distributionally}.}  
 
The next question is how to incorporate an explicit fairness criterion into this robustified formulation. The main concern about adding fairness constraints is that they must not break convexity; including a non-convex constraint to the problem would mean that standard convex solvers could no longer guarantee a global solution. This paper will, therefore, employ the separation penalty proposed by \citet{taskesen2020distributionally}. 
Specifically, we use their log-probabilistic equalised opportunities criterion, which leads to a convex relaxation of the equal opportunity criterion. We thus consider a solution maximally fair if $\Phi(\mathbb{Q}, \mathbf{w})$, defined as 
%
%
%
%
\begin{equation} \label{fairness_penalty}
\Phi(\mathbb{Q}, \mathbf{w}) = \Big| \mathbb{E_Q}\big[\ln\!\big(h(X;\mathbf{w})\big)\,|\,S\!=\!1,Y\!=\!1\big] 
- \mathbb{E_Q}\big[\ln\!\big(h(X;\mathbf{w})\big)\,|\,S\!=\!0,Y\!=\!1\big]\Big|,
\end{equation}
\noindent equals zero. In other words, 
a fair model should produce log-probabilities for positive (defaulter) outcomes that do not systematically differ between the protected and non-protected groups. 
 
Following \citet{taskesen2020distributionally}, we fix the marginal distributions of  $Y$ and $S$ to match those of the empirical distribution, ensuring that the ambiguity set perturbs only the conditional feature distribution while preserving the population proportions across outcome and sensitive groups. Imposing these marginal constraints prevents unrealistic distributional shifts that would distort the conditional expectations appearing in the fairness criterion, and, as \citet{wang2024wasserstein} also note, it enables a tractable reformulation in this setting. %
%
Denoting this constrained ambiguity set as 
\begin{equation}\label{constr_wasserst_ball}
\mathcal{B}_\rho^*(\mathbb{\hat{P}}_n)= \big\{\mathbb{Q}\in \mathcal{F}:\mathrm{W}(\mathbb{Q},\mathbb{\hat{P}}_n) \leq \rho, \; \mathbb{Q}(S\!=\!s, Y\!=\!y) = \hat{p}_{sy}, \; \forall s, y \in \{0,1\}  \big\},
\end{equation} 
where $\hat{p}_{sy} = \mathbb{\hat{P}}_n(S\!=\!s, Y\!=\!y)$, we can then put forward the following problem formulation: 
%
%
\begin{equation} \label{DRFLR}
\underset{\mathbf{w} \in \mathbb{R}^m}{\min}\; \underset{\mathbb{Q} \in \mathcal{B}_\rho^*(\mathbb{\hat{P}}_n) }{\sup}\; \mathbb{E}_\mathbb{Q}\big[\ell\big(Y,h(X;\mathbf{w})\big)\big] +\eta \, \Phi(\mathbb{Q}, \mathbf{w}).
\end{equation}
%
%

\noindent This objective function combines the log-loss term with a fairness penalty, making the formulation analogous to a form of regularised logistic regression, in which \(\eta\) determines how much emphasis is placed on fairness relative to accuracy. Both terms are evaluated under distributional uncertainty, so the resulting classifier is robust to deviations affecting performance or fairness. The parameters \( \kappa_s\) and \(\kappa_y\) defined earlier control how much individual observations of the sensitive attribute and outcome classes impact our defined distance (while preserving their marginal distributions). 

To simplify and shorten the notation, we will use the following names for the different classification methods included in this study:
\begin{itemize}
  \item LR: standard logistic regression \eqref{LR_emp_loss_min}, or, analogously, \eqref{DRFLR} with $\eta=0$ and $\rho=0$;\footnote{Note that for LR and LRL2, we utilise the definition in \eqref{LR_emp_loss_min} alongside the Newton-CG solver provided by scikit-learn, rather than CVXPY and MOSEK. Modelling these standard formulations in CVXPY introduces significant and unnecessary computational overhead. Because the underlying optimisation problems are convex, both approaches yield virtually identical results, with negligible variations due strictly to different solver convergence tolerances.}
  \item LRL2: standard regularised logistic regression \eqref{regul_log_reg} with an L2 penalty \eqref{ridge_pen};
  \item FLR: logistic regression with fairness penalty, i.e.\@ \eqref{DRFLR} with $\rho=0$, as setting the radius to zero collapses the Wasserstein ball to the empirical distribution;
  \item DRLR: Distributionally robust logistic regression \eqref{DRO_classifier}, that we have implemented as  \eqref{DRFLR} with $\eta=0$,
  so as to eliminate the fairness penalty;
  \item DRFLR: Distributionally robust fair logistic regression \eqref{DRFLR}.
  
\end{itemize}

By comparing DRFLR against these various models, we seek to gain further insight into the respective effects of either adding a fairness component (FLR), robustifying the model (DRLR), and, finally, combining both ideas (DRFLR), against our baseline models (LR and LRL2).
To fit these models, we adopt the relaxation described in the Appendix, and we solve the resulting optimisation problems using CVXPY with the MOSEK solver.
 
\noindent
\section{Data and performance evaluation}
\subsection{Data}
\label{data}
To evaluate the effectiveness of the proposed methods in a credit scoring setting, we have selected a number of credit scoring datasets that are publicly available and contain a suitable sensitive attribute, i.e.\@ age. The chosen datasets are: German Credit (GC) \citep{hofmann1994}; Give Me Some Credit (GMSC) \citep{GiveMeSomeCredit}; Home Credit (HC) \citep{homecr}; Taiwan Credit \citep{i-cheng_2009}; and Pacific-Asia Knowledge Discovery and Data mining conference (PAKDD) \citep{ealigam}. Due to computational constraints and the lack of parallelisation capability, all but GC were subsampled to 5000 entries, while preserving the marginal distributions of the default variable and the sensitive attribute. It is worth noting that some of these datasets (e.g.\@ German Credit) also include gender (encoded within the marital status); however, for simplicity and consistency of the comparisons, we have, in each case, selected age as our sole sensitive attribute. For a more detailed description of each dataset, we refer the reader to Table~\ref{table1}.

\begin{table}[h]
\caption{List of datasets and description \\ \scriptsize *Subsampled to 5000 observations while preserving the marginal distributions of $y$ and $s$} 
\label{table1}
\scriptsize

    \begin{tabularx}{\textwidth}{XXp{4.8cm}X} 
     \toprule
     Name & Description  & Marginal  distribution& Source \\  
     \hline
     German Credit (GC) & 13 categorical features, 7 numerical features, 1K observations 
     & $y$=0,\;$s$=0: 0.5900; $y$=0,\;$s$=1: 0.1100; \newline $y$=1,\;$s$=0: 0.2200; $y$=1,\;$s$=1: 0.0800 &  \citep{hofmann1994} available at the UCI ML Repository \citep{Dua:2019} \\ 
     \hline
     Give me some credit (GMSC) & 12 numerical features, 150K observations* & $y$=0,\;$s$=0: 0.9154; $y$=0,\;$s$=1: 0.0179; \newline $y$=1,\;$s$=0: 0.0645; $y$=1,\;$s$=1: 0.0022 &  \citep{GiveMeSomeCredit} available  for Kaggle competitions\\
     \hline
     Home Credit (HC) & 122 numerical features, 308K observations* & $y$=0,\;$s$=0: 0.8848; $y$=0,\;$s$=1: 0.0346; \newline $y$=1,\;$s$=0: 0.0758; $y$=1,\;$s$=1: 0.0048 & \citep{homecr} available  for Kaggle competitions \\
      \hline
     Taiwan Credit (TC) & 24 numerical features, 30K observations* & $y$=0,\;$s$=0: 0.6842; $y$=0,\;$s$=1: 0.0946; \newline $y$=1,\;$s$=0: 0.1868; $y$=1,\;$s$=1: 0.0344  &  \citep{i-cheng_2009} available  at the UCI ML Repository \citep{Dua:2019}\\
     \hline
     Pacific-Asia Knowledge Discovery and Data Mining Conference (PAKDD) & 52 numerical features, 50K observations* &$y$=0,\;$s$=0: 0.6637; $y$=0,\;$s$=1: 0.0755; \newline $y$=1,\;$s$=0: 0.2214; $y$=1,\;$s$=1: 0.0394 &  \citep{ealigam} available for competitions \\
     \bottomrule
    \end{tabularx}
\normalsize
\end{table}

Some basic pre-processing was applied to each dataset, to facilitate the classifier comparison. Specifically, we one-hot encoded each of the categorical features, keeping the ten most common categories in the few instances where a feature had a large number of categories. We also substituted missing values with the median for continuous features and kept them as a separate category for categorical features. 
Lastly, to reduce dimensionality, we have applied Lasso feature selection, with the tuning parameter set to a value between 0.01 and 0.001\footnote{0.01 for GC and GMSC; 0.005 for HC; and 0.001 for TC \& PAKDD.} \citep{tibshirani1996regression}.

\subsection{Performance evaluation}
To compare the classifiers, we will use the area under the Receiver Operating Characteristic curve (ROC), as this is a standard method of measuring discriminative performance in credit scoring. For all threshold (cut-off) values $\tau \in [0,1]$, the ROC curve plots the True Positive Rate (TPR) (also called sensitivity) against the False Positive Rate (FPR) (or $1 - \text{specificity}$),  i.e. 
\begin{equation} \label{ROC}
 \mathrm{TPR}(\tau) = \frac{1}{N_1} \sum_{i: y_i=1} \mathbb{I}(\hat{y}_i \ge \tau), \quad 
 \mathrm{FPR}(\tau) = \frac{1}{N_0} \sum_{i: y_i=0} \mathbb{I}(\hat{y}_i \ge \tau), 
\end{equation}
where $\mathbb{I}(\cdot)$ is the indicator function that equals 1 if the condition is true, or 0 otherwise, and $N_{1}$ and $N_{0}$ are the number of observations for which $y = 1$ and $y = 0$, respectively. 

To evaluate fairness, we will use a degree 
of separation measure (SP) used by \citet{kozodoi2022fairness} and based on a criterion proposed by \citet{NIPS2016_9d268236}. 
This measure considers the difference in the false positive and true positive rates 
between the protected and non-protected group (smaller values implying greater fairness), as follows:
\small
    \begin{equation} \label{eq:17}
    \mathrm{SP} = \dfrac{1}{2} \,\big|\,(\mathrm{TPR}_{s=1}-\mathrm{TPR}_{s=0})+(\mathrm{FPR}_{s=1}-\mathrm{FPR}_{s=0})\,\big|, \end{equation}    
\noindent
\normalsize
with $\mathrm{TPR}_{s=s^*}(\tau) = \frac{1}{N_{1, s^*}} \sum_{i: y_i=1, s_i=s^*} \mathbb{I}(\hat{y}_i \ge \tau)$, $\mathrm{FPR}_{s=s^*}(\tau) = \frac{1}{N_{0, s^*}} \sum_{i: y_i=0, s_i=s^*} \mathbb{I}(\hat{y}_i \ge \tau)$,
and letting $N_{y^*\!, s^*}$ denote the number of observations having values $y^{*}$, $s^*$ for $y$ and $s$, respectively. 

\begin{figure}[!htb] 
    \centering 
    \includegraphics[width=0.6\imagewidth]{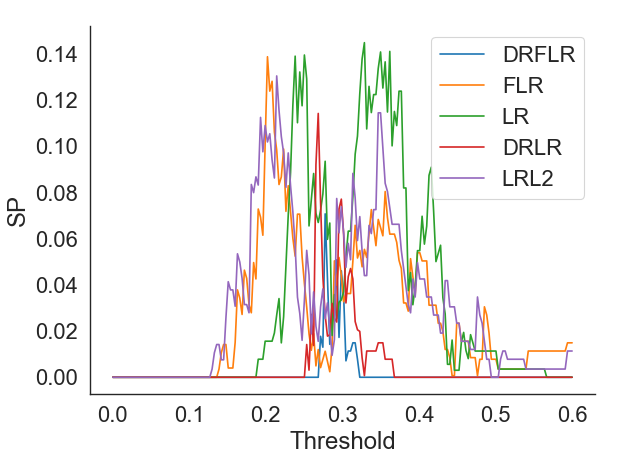}
    \caption{Sensitivity of SP to changes in application threshold; GC example; 
    $\rho=0.01$; $\eta=0.04$; $\kappa_y= \kappa_s=0.2$}
    \label{fig:1}
    \end{figure}

However, this measure has a significant drawback for credit scoring in that it depends on the classification threshold and, thus, the lender's risk appetite. As such, it provides limited information on the fairness of the estimated PD further away from the cut-off score. Using the models for the GC dataset as an example, Figure \ref{fig:1} illustrates that conclusions based on SP about which model is fairest heavily depend on the chosen threshold.

For this reason, we also report a second fairness metric, Log Probabilistic Equalised Opportunities (LEO). This is similar to the measure used by the DRFLR system to penalise unfairness, but now applied to our sample. Unlike SP, it does not depend on a chosen threshold, but instead considers the absolute difference in the expected value (mean) of the log PD of the defaulters between the protected 
and non-protected 
subgroup, as follows: 
%
    \begin{equation} 
    \label{LEO}
\mathrm{LEO} =
\Big|\,
\mathbb{E}_{\hat{\mathbb{P}}_n}
\big[ \ln \hat{y} \;\big|\; s\!=\!1,\,y\!=\!1 \big]
-
\mathbb{E}_{\hat{\mathbb{P}}_n}
\big[ \ln \hat{y} \;\big|\; s\!=\!0,\,y\!=\!1 \big]
\,\Big|.
    \end{equation}
\noindent Note that this measure focuses on 
defaults ($y$=1) only
, giving us an indication of how much the risk estimates tend to differ between the two subgroups of interest among those.   
 
A potential drawback of LEO is that it can be over-optimistic with regularised classifiers as those tend to produce narrower PD distributions, thus reducing any such differences. Hence, we will compare model fairness using both LEO and SP. To evaluate SP, we select a threshold that maximises the difference in the true positive rate and false positive rate; this difference is commonly known as the Youden J statistic \citep{youden1950index} and when measured at the threshold that returns its maximum, it is known as the Kolmogorov–Smirnov statistic \citep{kolmogorov1933, smirnov1948table}.
 
We will use five-fold cross-validation to evaluate model performance and fairness; hence, each of the reported metrics is averaged over five model runs.

\section{Results}
This section presents the results of the experiments applying the five classifiers to the five credit datasets listed above. As DRFLR introduces distinct, specialised hyperparameters, we include experiments for each of these and explain their effects, 
to guide credit scorers aiming to use DRO.  

\subsection{Hyperparameter tuning}

\begin{figure}[h]
    \centering
    
    \begin{subfigure}[b]{\multilen}
        \centering
        \includegraphics[trim=0 0.86cm 0 0, clip,width=0.6\imagewidth]{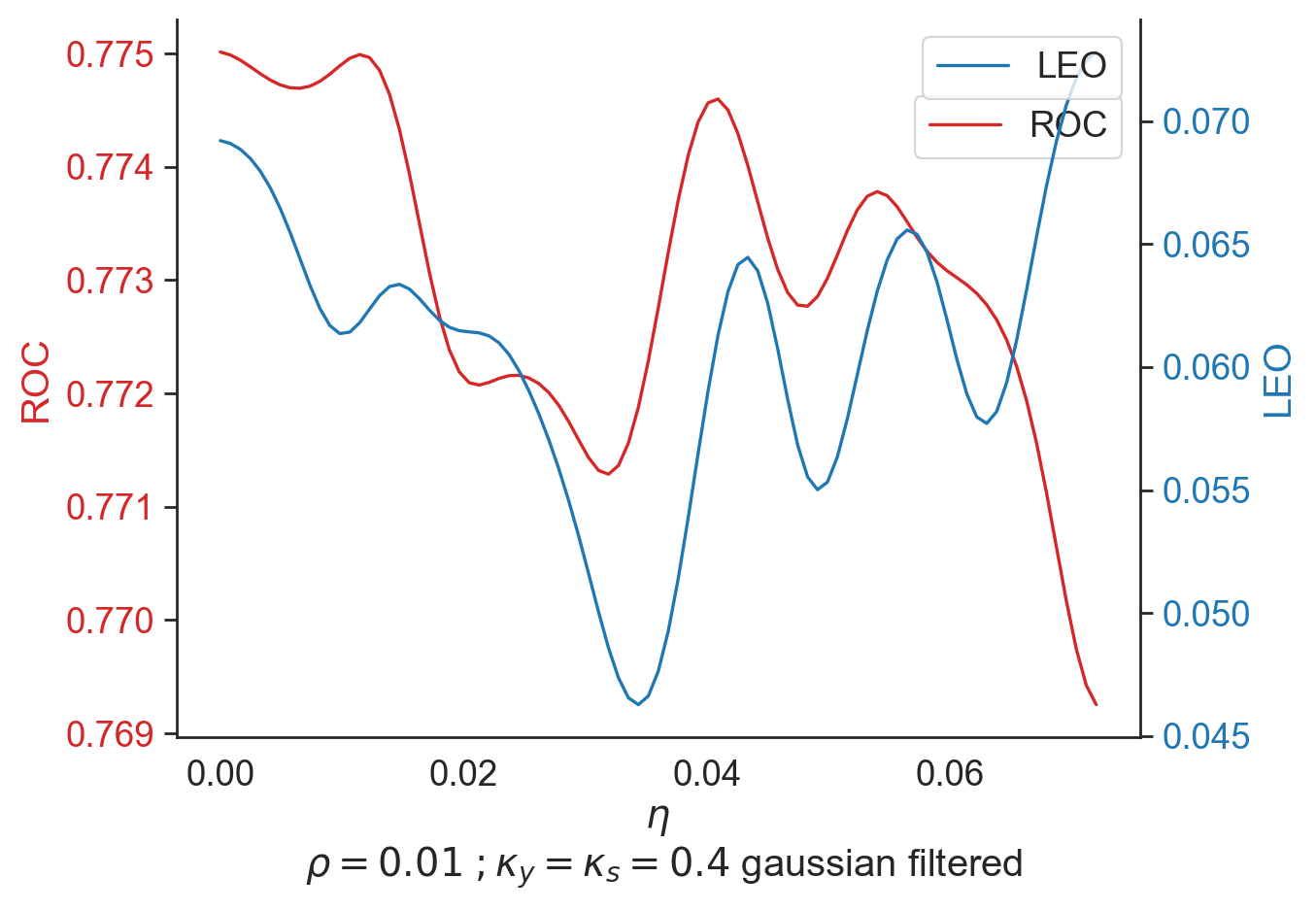}
        \caption{Effect of changes in $\eta$}
        \label{fig:2a}
        \vspace{2pt} 
        {\centering\scriptsize \(\rho = 0.01 ;\, \kappa_y=\kappa_s=0.4\) gaussian filtered\par}
    \end{subfigure}
   \hfill
    \begin{subfigure}[b]{\multilen}
        \centering
        \includegraphics[trim=0 0.86cm 0 0, clip, width=0.6\imagewidth]{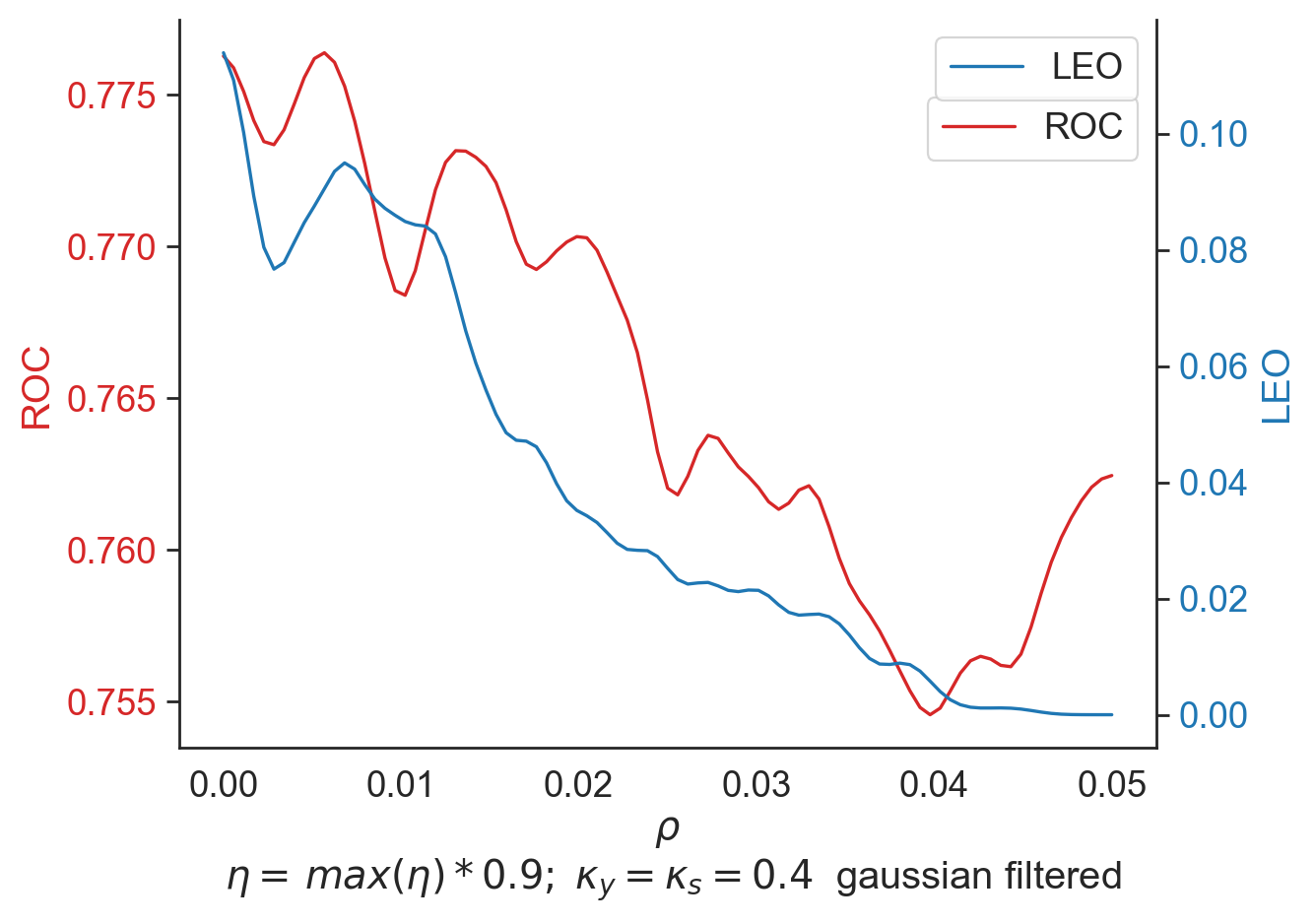}
        \caption{Effect of changes in $\rho$}
        \label{fig:2b}
        \vspace{2pt} 
        {\centering\scriptsize \(\eta = \eta_{max}*0.9 ;\, \kappa_y=\kappa_s=0.4\) gaussian filtered\par}
    \end{subfigure}
    
    \caption{Evolution of ROC and LEO with changes in $\eta$ and $\rho$ (GC)}
    \label{fig:rhoeta}
\end{figure}

To arrive at a tractable formulation,  \cite{taskesen2020distributionally} require that the fairness penalty weight $\eta$ be kept bounded between 0 and $\min(\hat{p}_{01}, \hat{p}_{11})$, where, as in section \ref{methods}, 
$\hat{p}_{sy}$ denotes the empirical proportion of applicants with values $s$ and $y$ for the sensitive and outcome attribute, respectively.  
Given the high level of class imbalance in the datasets (which is often the norm in credit scoring), this upper bound value remains fairly small, the largest being for GC with $\eta_{ \max } \approx 0.08$ and the lowest for GMSC with $\eta_{ \max }  \approx 0.001$. Even in the case of GC (see Figure \ref{fig:2a})\footnote{To highlight 
the underlying performance trends, the plotted curves were smoothed using a one-dimensional Gaussian filter from SciPy's \texttt{ndimage} module \citep{2020SciPy-NMeth}, with a standard deviation 
of 2. This technique applies a Gaussian-weighted moving average to the metric sequences, suppressing high-frequency experimental noise to make the macroscopic trends visually interpretable without shifting peak locations.},
though, we can observe how, possibly due to this limited range, the values for ROC and LEO fail to show a clear downward trend as $\eta$ is increased. 

Since our Wasserstein distance calculations incorporate the sensitive attribute, it can be expected that the tuning of the Wasserstein ball radius ($\rho$) has some effect on fairness. However, care should be taken when setting this parameter, as predictive performance can degrade drastically with larger values. It is worth noting here that our grid search was configured to look for optimal ROC performance. Therefore, the resulting $\rho$ will tend to be small, as any distributional differences between the random cross-validation splits are likely to be limited. Since, in an operational environment, we can reasonably expect that the data fed to the deployed model will be more different from the training data, we suggest that the final $\rho$ be set to a slightly higher value than the one found during the grid search when the aim would be to deploy the models in a real-life setting.

\begin{figure}[h]
       \centering
        \includegraphics[trim=0 0.87cm 0 0, clip, width=0.6\imagewidth]{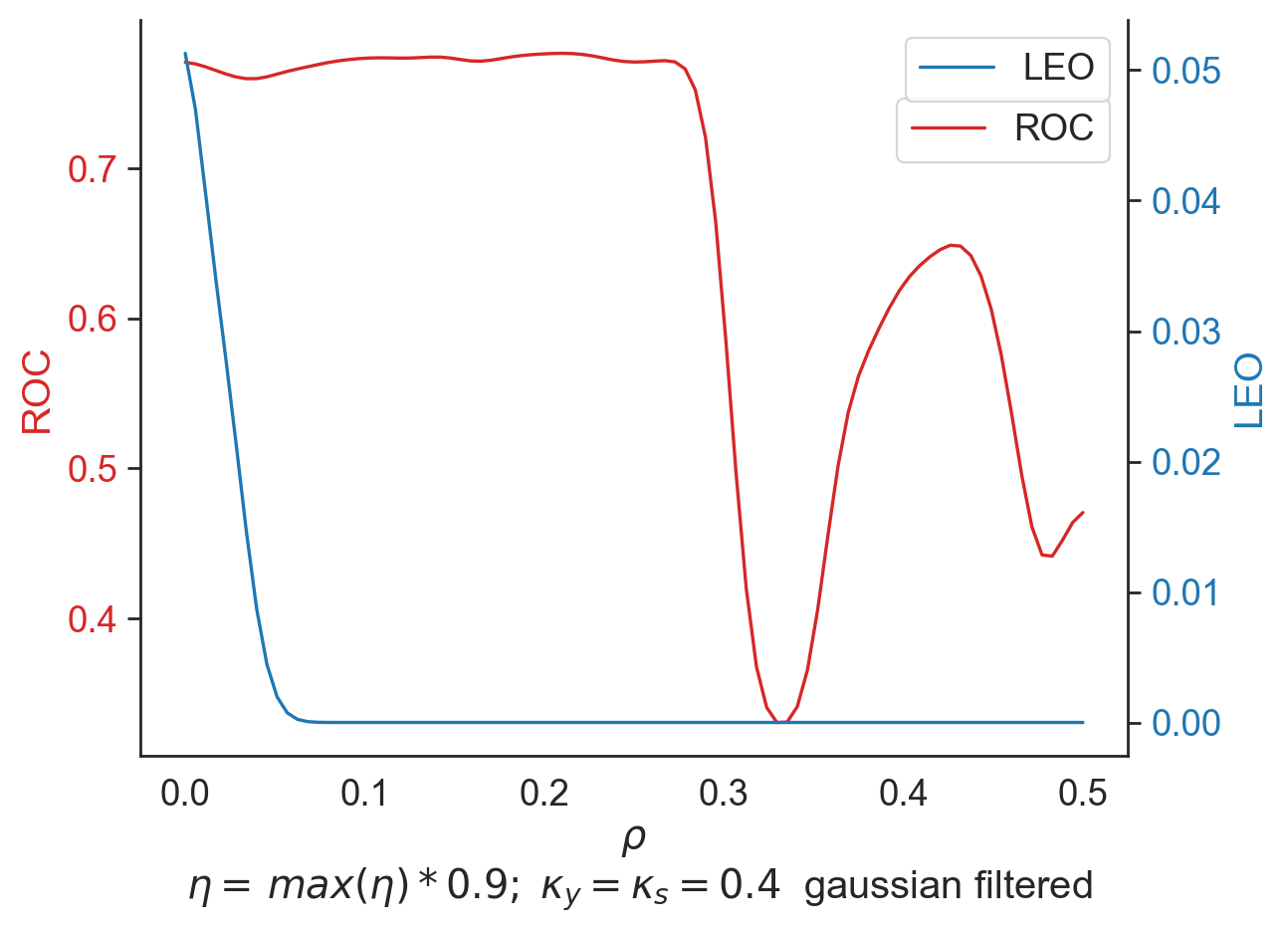}
        
        \caption{Evolution of ROC and LEO over a broader span of $\rho$ values (GC)}
        \label{appendix:3}
        
        \vspace{2pt} 
        {\centering\scriptsize \(\eta = \eta_{max}*0.9 ;\, \kappa_y=\kappa_s=0.4\) gaussian filtered\par}
\end{figure}

As expected, Figure \ref{fig:2b} shows how small changes in $\rho$ can indeed have a large impact on both LEO and ROC. Interestingly, the trade-off is such that good performance can still be achieved with near-zero LEO. 
However, predictive performance tends to drop sharply once $\rho$ further exceeds some critical value, for example, from around 0.3 in the case of GC (see Figure \ref{appendix:3}). 

\begin{figure}[h]

  \begin{subfigure}[b]{\multilen}
    \includegraphics[width=0.60\imagewidth]{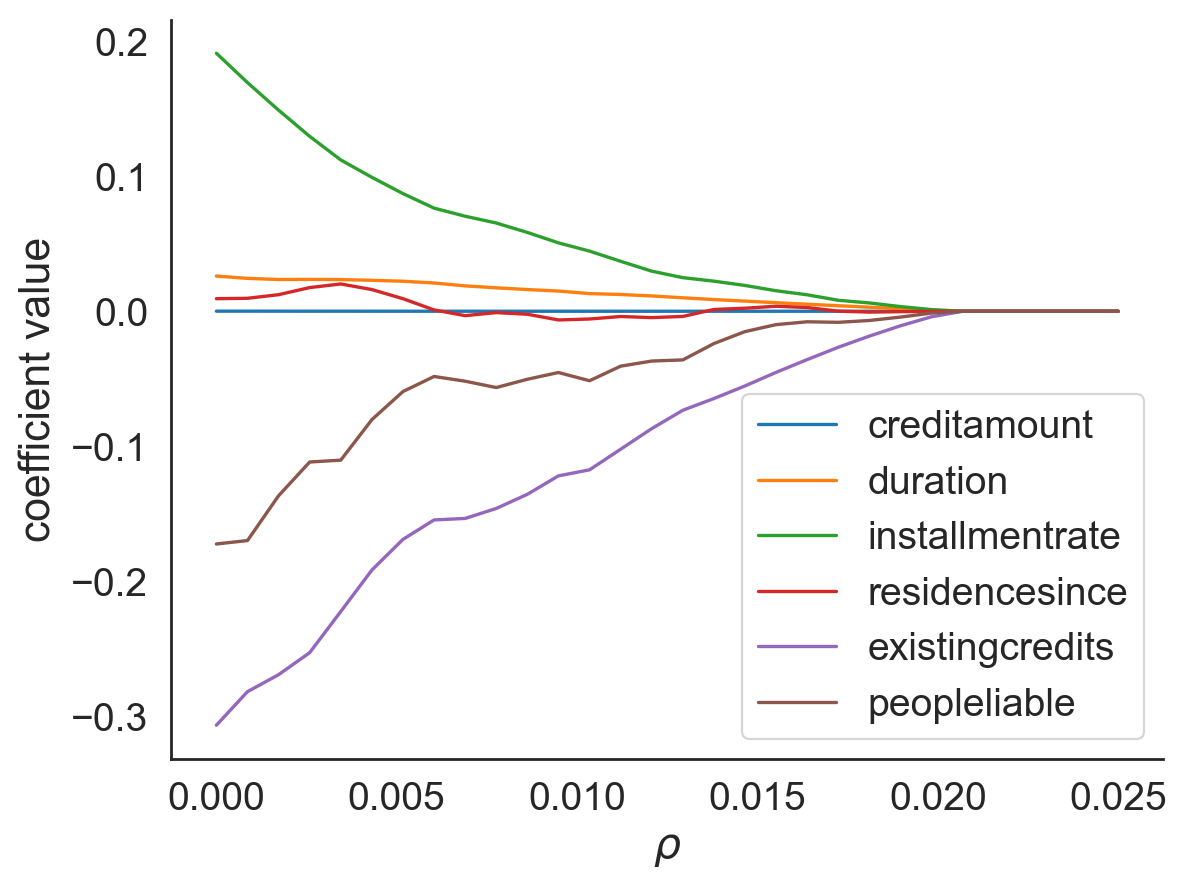} 
    \caption{Evolution of most relevant coefficients } 
    \label{appendix:coef}
  \end{subfigure}
  \hfill
  \begin{subfigure}[b]{\multilen}
    \includegraphics[width=0.60\imagewidth]{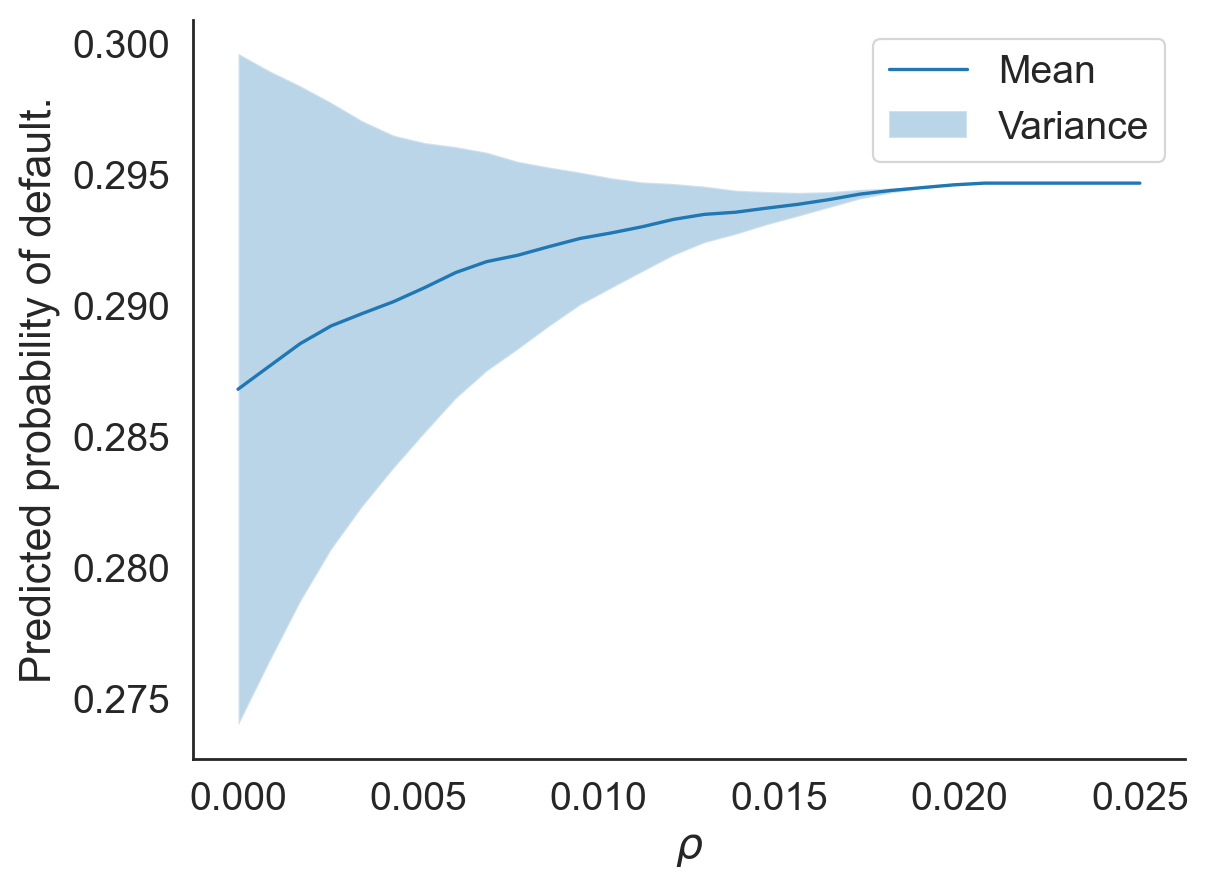}
    \caption{Evolution of the expected PD  } 
    \label{appendix:probas}
  \end{subfigure} 
\caption{Evolution of coefficients and PD distribution as a function of $\rho$ (GC; $\kappa_y= \kappa_s=0.4;\;\eta=0.05$)}
        
 \end{figure}

As mentioned in section \ref{methods}, DRO can be viewed as a regularisation method, albeit one that does not directly impose structure on the parameter space, as L1 or L2 regularisation do. Indeed, increasing $\rho$ induces coefficient shrinkage, consistent with a regularisation effect (as seen in Figure \ref{appendix:coef}), thus yielding a narrower distribution of predicted PD (Figure \ref{appendix:probas}). As explained in section \ref{data}, this can have had a direct bearing on our reported fairness metric, LEO.

\begin{figure}[htb]
    \centering
    \begin{subfigure}[b]{\multilen}
        \centering
        \includegraphics[width=0.55 \imagewidth]{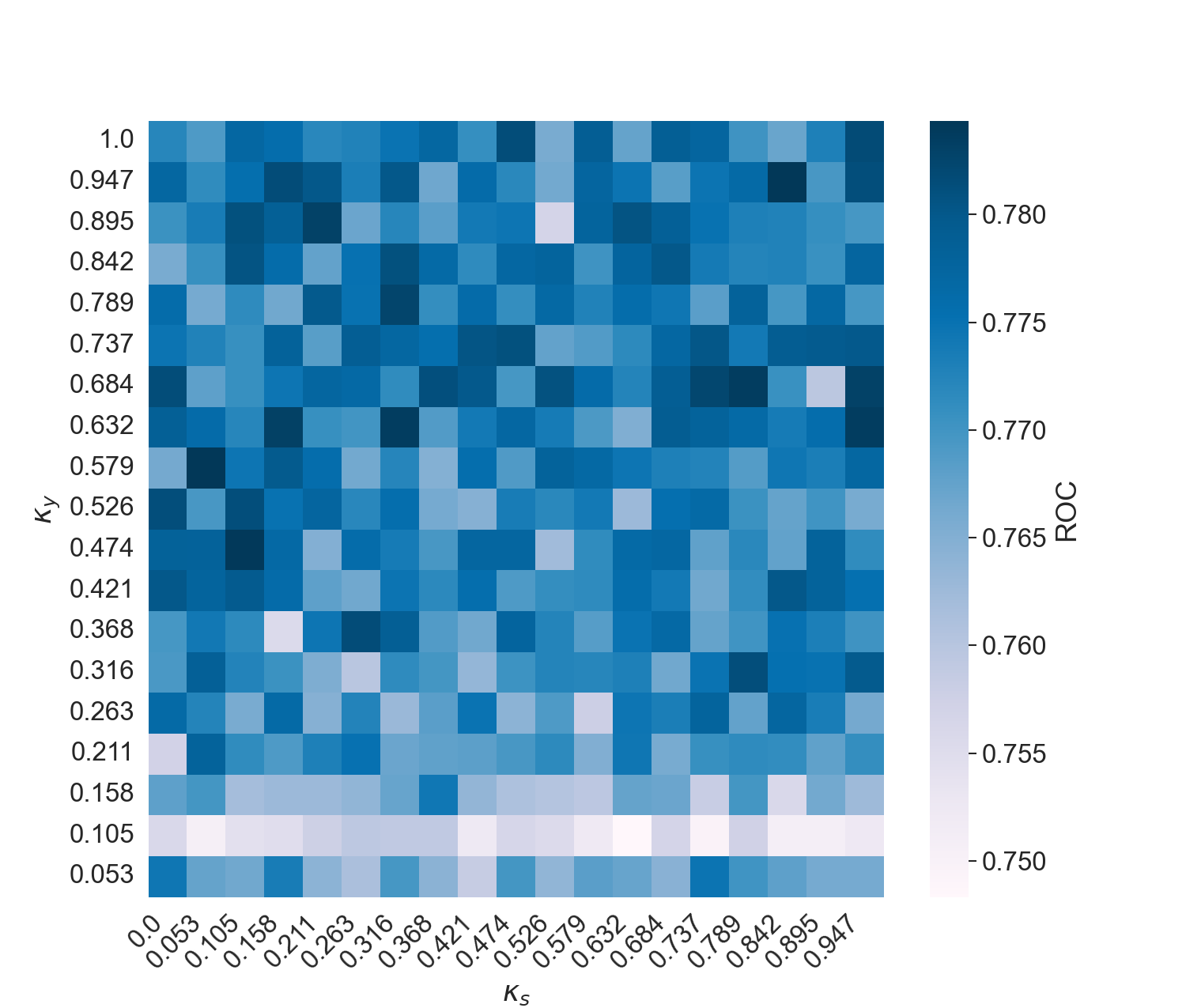}
       
        \caption{ROC}
        \label{fig:sub1}
    \end{subfigure}
   \hfill
    \begin{subfigure}[b]{\multilen}
        \centering
        \includegraphics[width=0.55\imagewidth]{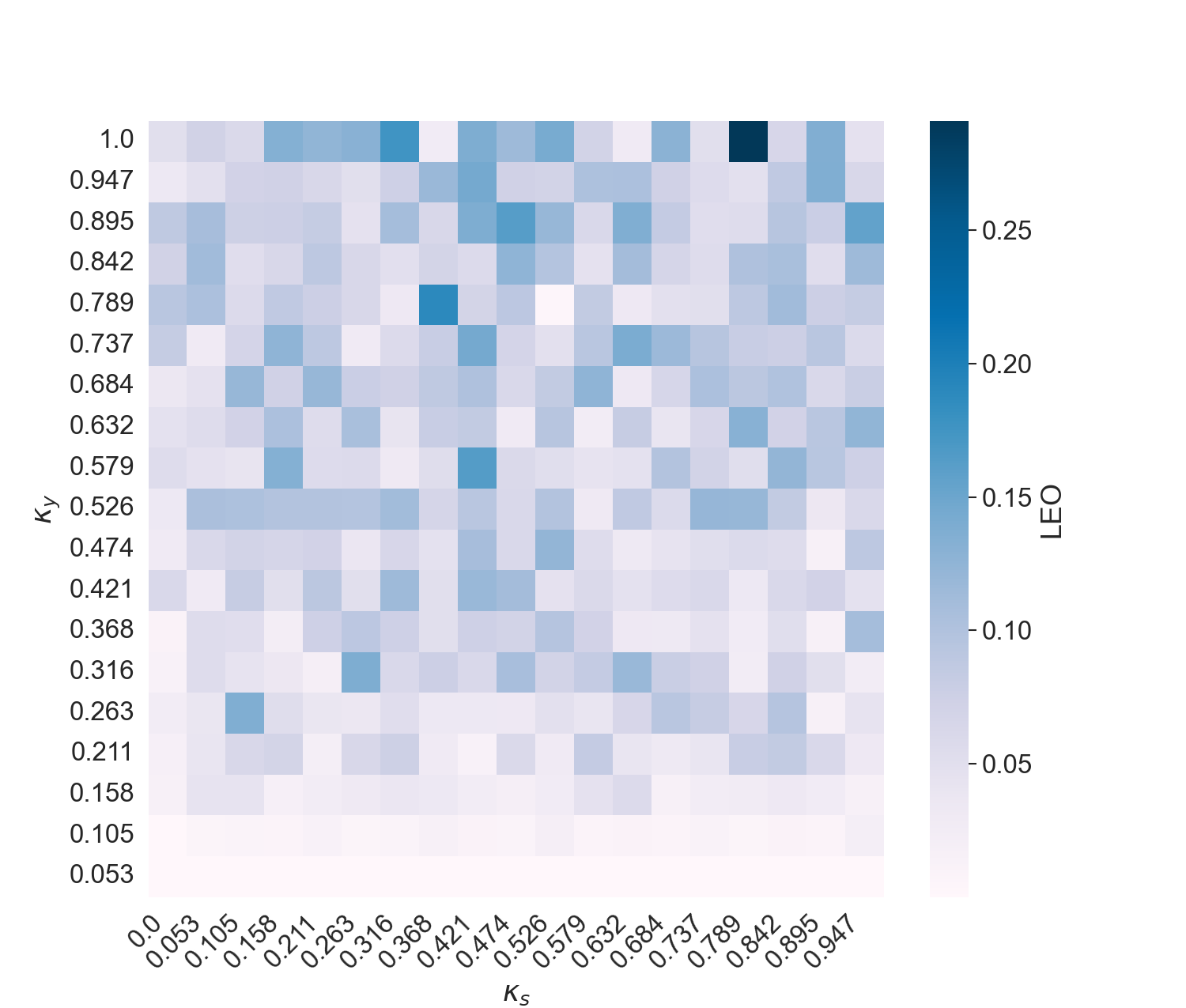}
        
        \caption{LEO}
        \label{fig:sub2}
    \end{subfigure}
    
    \caption{Evolution of ROC and LEO as a function of $\kappa_y$ and $\kappa_s$ (GC; $\rho=0.01;\;\eta=0.05$)}
    \label{fig:3}
    
\end{figure}

The remaining two hyperparameters to consider are $\kappa_y$ and $\kappa_s$. No discernible effect of tuning $\kappa_s$ is observed in Figure \ref{fig:3}. However, it is worth noting that very low values of $\kappa_y$ yield the best results in terms of LEO, at the expense of lower ROC.\footnote{We have omitted any result values for $\kappa_y=0$ as the ROC of around 0.5 associated with it would make the colour gradient harder to differentiate elsewhere on the map.} 

This section has focused on exploring these four hyperparameters as they differ from those known to most ML practitioners, and the provided analyses help explain the results provided in the following section. For the remainder of the experiments, we have employed grid search to tune the  $\rho$ and $\kappa$ parameters with respect to ROC performance, letting $\rho$ range from $0.01$ to $0.05$, 
and $\kappa_y$ and $\kappa_s$ from $0.2$ to $0.5$. To assess the extent to which adding the fairness penalty affects findings, we kept $\eta$ fixed to $0.9\times \eta_{\max} $
, since 
tuning this parameter to optimise predictive performance would often return its smallest value in the grid search range. 

\subsection{Performance comparison}
Table \ref{table:perf} summarises how each of the methods performs on the five sample datasets, with respect to our three criteria: ROC, LEO, and SP. 
\vspace{+4pt}
\begin{table*}[!h]
\centering
\caption{Performance comparison of the selected classifiers on different credit scoring data samples \\ 
\scriptsize All hyperparameters tuned using grid search, except for $\eta$, which was fixed to $0.9\times\eta_{max}$.
Values below 0.0005 are displayed in scientific notation, where ‘e’ denotes $ \times10 $. Best results for each metric and model are displayed in bold (apparent ties are resolved by evaluating unrounded values at higher decimal precision), and $\pm$ refers to the standard deviation.}
\centering

\scriptsize
\begin{tabular}{llccccc}

\toprule

\multirow{2}{*}{Metric} & \multirow{2}{*}{Model} & \multicolumn{5}{c}{Dataset} \\

& & GC & HC & TC & PAKDD & GMSC \\
\midrule
\multirow{5}{*}{ROC}   &
LR & $0.763 \pm 0.052$   &  
$0.722 \pm 0.036$ &  
$0.701 \pm 0.032$   &  
$0.585 \pm 0.02$ &  
$0.642 \pm 0.046$ \\

&LRL2 &  $\textbf{0.765} \pm \textbf{0.052}$ & 
$0.723 \pm 0.036$ &   
$\mathbf{0.701 \pm 0.031}$  &
$0.585 \pm 0.02$ &  
$0.641 \pm 0.046$ \\

&FLR & $0.764 \pm 0.052$ &
$0.721 \pm 0.036$ &
$0.699 \pm 0.034$ &
$0.585 \pm 0.02$ & 
$0.639 \pm 0.043$ \\

&DRLR & $0.759 \pm 0.051$ &    
$\mathbf{0.723\pm 0.032}$ &   
$0.700 \pm 0.034$ &   
$0.587\pm 0.021$ &  
$\mathbf{0.723 \pm 0.046}$ \\
& 
DRFLR & $0.759 \pm 0.051$ & 
$0.721 \pm 0.030$ &  
$0.700 \pm 0.034$ &   
$\mathbf{0.587 \pm 0.021}$ & 
$0.710 \pm 0.035$ \\

\midrule
\multirow{5}{*}{LEO} &

LR &  $0.219 \pm 0.169$  &  
 $0.778 \pm 0.187$ &  
 $0.197 \pm 0.140$ &     
 $0.055 \pm 0.067$ &      
$0.338 \pm 0.248$ \\

& LRL2 & $0.211 \pm 0.166$ &     
$0.654 \pm 0.13$&     
$0.196 \pm 0.139$ &    
$0.045 \pm 0.066$ &  
$0.321 \pm 0.255$ \\

& FLR &  $0.204 \pm 0.171$&  

$0.737 \pm 0.210$ &
 $0.197\pm 0.144$  &
$0.053 \pm 0.064$ &
$0.332 \pm 0.269$\\

& DRLR &  $0.117\pm 0.133$ & 
 $0.01 \pm 0.002$  & 
 $0.004 \pm 0.011$  & 
 $\mathbf{0.002 \pm 0.004}$ & 
$2.46e^{-05} \pm 5.17e^{-05}$\\

& DRFLR &  $\mathbf{0.088 \pm 0.117}$ & 

$\mathbf{0.002 \pm 0.005}$ & 
$\mathbf{0.001\pm 0.003}$   &
 $0.002 \pm 0.004$ & 
 $\mathbf{1.65e^{-06} \pm 2.76e^{-06}}$ \\

\midrule
\multirow{5}{*}{SP} & 

LR & $8.47e^{-04} \pm 0.003$  &      
$0.007 \pm 0.011$ &  
$0.006 \pm 0.019$  & 
$0.007 \pm 0.006$ &  
$0.086 \pm 0.049$\\

& LRL2 & ${8.47e^{-04} \pm 0.003}$ &     
$0.003 \pm 0.005$  &    
$0.01 \pm 0.01$ &  
$\mathbf{0.007\pm 0.006}$ &
$0.094 \pm 0.052$\\

& FLR & $\mathbf{0\pm 0}$ &    
$\mathbf{3.39e^{-04} \pm 3.10e^{-04}}$&
 $0.006 \pm 0.011$ &
$0.011\pm 0.012$ & 
$0.083 \pm 0.056$\\

& DRLR & $0.005 \pm 0.014$ &    
$0.003 \pm 0.006$& 
$0.007 \pm 0.01$ & 
$0.015 \pm 0.017$&  
$0.01 \pm 0.011$\\

& DRFLR & $0.004\pm 0.012$ & 
$0.007 \pm 0.01$ &
$\mathbf{0.003 \pm 0.006}$ &
$0.011\pm 0.019$ &
$\mathbf{0.01\pm 0.011}$ \\

\bottomrule
\end{tabular}

\label{table:perf}
\end{table*}


In Table \ref{table:perf}, we can observe how DRFLR, on four out of five datasets, provides the smallest LEO values (indicating fairest treatment), with minimal loss in ROC performance. In most scenarios, the close second-best performer in terms of LEO is DRLR, which could be explained by an improvement in generalisation through robustification or the tendency of LEO to favour regularised methods (as explained in section \ref{data}). It is important to note, however, that our ground metric considers changes in the sensitive attribute when identifying the worst-case distribution; hence, our DRLR formulation may indirectly affect fairness even though it does not include an explicit fairness penalty. 

In contrast, with regard to SP, no clear winner emerges. This illustrates how different fairness criteria may favour different model solutions. Hence, unless the model is explicitly tuned for this criterion, the proposed distributionally robust methods are unlikely to have a consistent effect on SP.

Finally, across most datasets (GC, HC, TC, and PAKDD), the differences in ROC are marginal and well within the standard deviation margins, making them too close to call a definitive winner in terms of predictive performance. However, GMSC is a notable exception: here, the robust methods (DRLR and DRFLR) show a distinct performance lift, outperforming the standard logistic regression by roughly 0.07–0.08 in ROC. 
This is likely due to the large distributional feature distances among the subgroups observed in this dataset (see Table \ref{tab:wasserstein_meaningful_compact}). 

\begin{table*}[!h]
\centering
\caption{Pairwise feature space heterogeneity evaluation across subgroups \\
\scriptsize Table values show the mean feature-wise Wasserstein distance between sub-populations defined by $(S\!=\!s,Y\!=\!y)$ in the original datasets. Binary labels $sy$ identify the subgroups, with the first digit denoting $S$ and the second $Y$. For example, $00\!\leftrightarrow\!01$ denotes the distance between $(S\!=\!0,Y\!=\!0)$ and $(S\!=\!0,Y\!=\!1)$, whereas $s0\! \leftrightarrow \!s1$ compares the subgroups with $Y\!=\!0$ and $Y\!=\!1$, respectively, marginalising over $S$. Maximum values are shown in bold.}

\label{tab:wasserstein_meaningful_compact}
\setlength{\tabcolsep}{6pt} 
\scriptsize
\begin{tabularx}{\textwidth}{lXXXXXXXX}
\toprule
\textbf{Dataset} & 
  $0y \!\! \leftrightarrow \!\! 1y$ & 
  $s0 \!\! \leftrightarrow \!\! s1$ & 
  $00 \!\! \leftrightarrow \!\! 01$ & 
  $10 \!\! \leftrightarrow \!\! 11$ & 
  $00 \!\! \leftrightarrow \!\! 10$ & 
  $01 \!\! \leftrightarrow \!\! 11$ & 
  $01 \!\! \leftrightarrow \!\! 10$ & 
  $00 \!\! \leftrightarrow \!\! 11$ \\
\midrule
German Credit & 0.2723 & \textbf{1.1295} & \textbf{1.1373} & 1.1824 & 0.4548 & 0.2763 & \textbf{1.3182} & 1.0125 \\
GMSC & \textbf{0.6043} & 0.4945 & 0.4523 & \textbf{1.5442} & \textbf{0.4688} & \textbf{1.6348} & 0.3426 & \textbf{1.9875} \\
Taiwan Credit & 0.2876 & 0.6323 & 0.6129 & 0.6860 & 0.2365 & 0.3098 & 0.5272 & 0.9221 \\
Home Credit & 0.4585 & 0.3441 & 0.3361 & 0.3539 & 0.4556 & 0.3892 & 0.2541 & 0.6845 \\
PAKDD & 0.2088 & 0.0927 & 0.0954 & 0.0471 & 0.2243 & 0.1790 & 0.1897 & 0.2276 \\
\bottomrule
\end{tabularx}
\end{table*}

These results demonstrate that robust methods can, in most cases, be expected to perform on par with standard models, while yielding substantial performance gains on specific datasets. Moreover, they suggest that the aforementioned reductions in LEO can be achieved without compromising predictive performance. 

\subsection{Changes in marginal proportions}
\begin{figure}[h]

  \begin{subfigure}[b]{\multilen}
    \includegraphics[trim={0 0 0 0},clip,width=0.6\imagewidth]{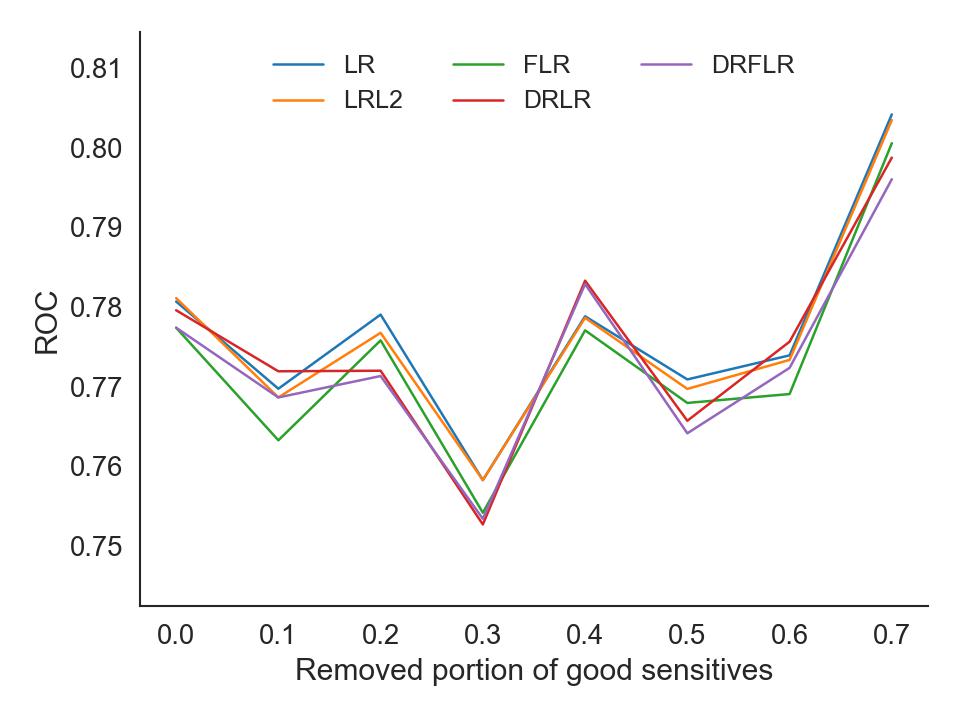}
    \caption{ROC} 
  \end{subfigure} 
  \begin{subfigure}[b]{\multilen}
    \includegraphics[trim={0 0 0 0},clip,width=0.6\imagewidth]{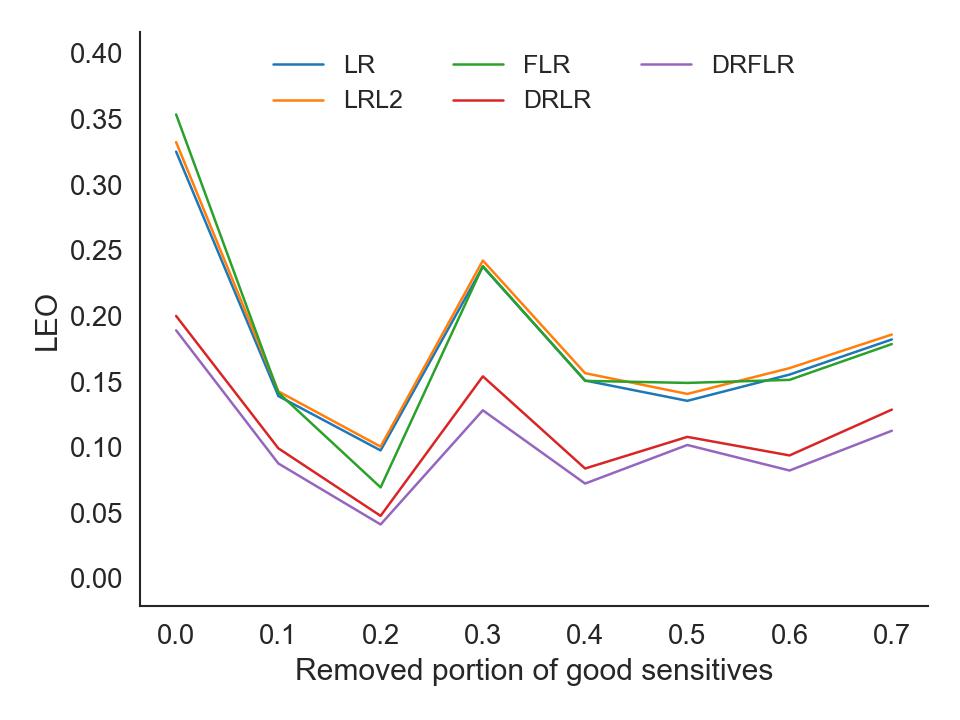} 
    \caption{LEO} 
  \end{subfigure} 
  \begin{center}
   \begin{subfigure}[b]{\multilen}
    \includegraphics[trim={0 0 0 0},clip,width=0.6\imagewidth]{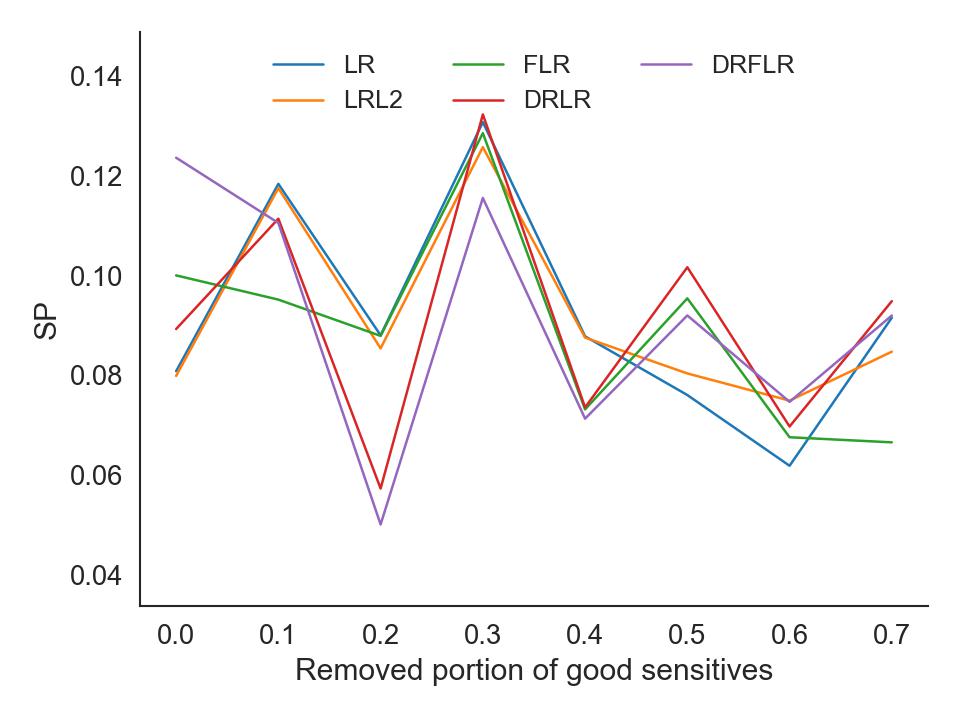} 
    \caption{SP} 
  \end{subfigure} 

\caption{Evolution of ROC, LEO, and SP with changes in marginal distribution (GC)}
\label{fig:4} 
    \text{\footnotesize $\rho=0.01$, $\kappa_y=\kappa_s=0.4$, $\eta=\dfrac{\eta_{max}}{1.5}$}
\end{center}
\end{figure}

To test how sensitive our findings are to distributional properties such as the level of class imbalance, we consider a scenario where the training data would contain fewer examples of good payers from the protected group. Specifically, for GC, Figure \ref{fig:4} shows the impact on ROC, LEO, and SP, of randomly removing a growing proportion of non-defaulting protected-group applicants. This might mirror a setting where, historically, fewer such applicants applied or were previously accepted by the lender. 


\begin{figure}[h] 
  \centering
  \begin{subfigure}[b]{\multilen}
    \includegraphics[width=0.54\imagewidth]{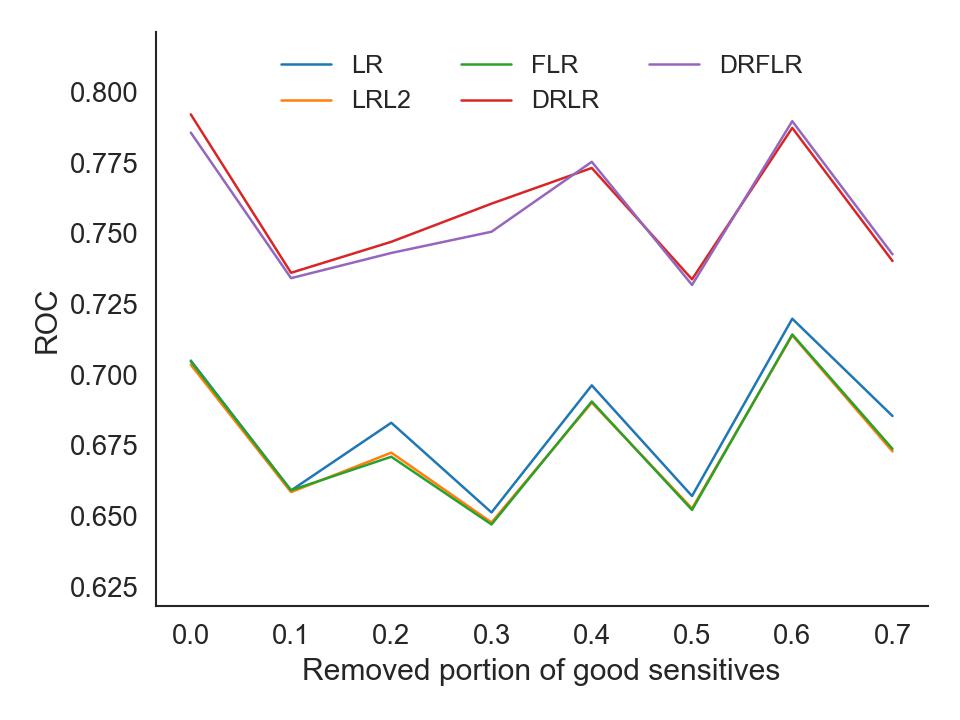}
    \caption{GMSC} 
  \end{subfigure}
  \hfill
  \begin{subfigure}[b]{\multilen}
    \includegraphics[width=0.54\imagewidth]{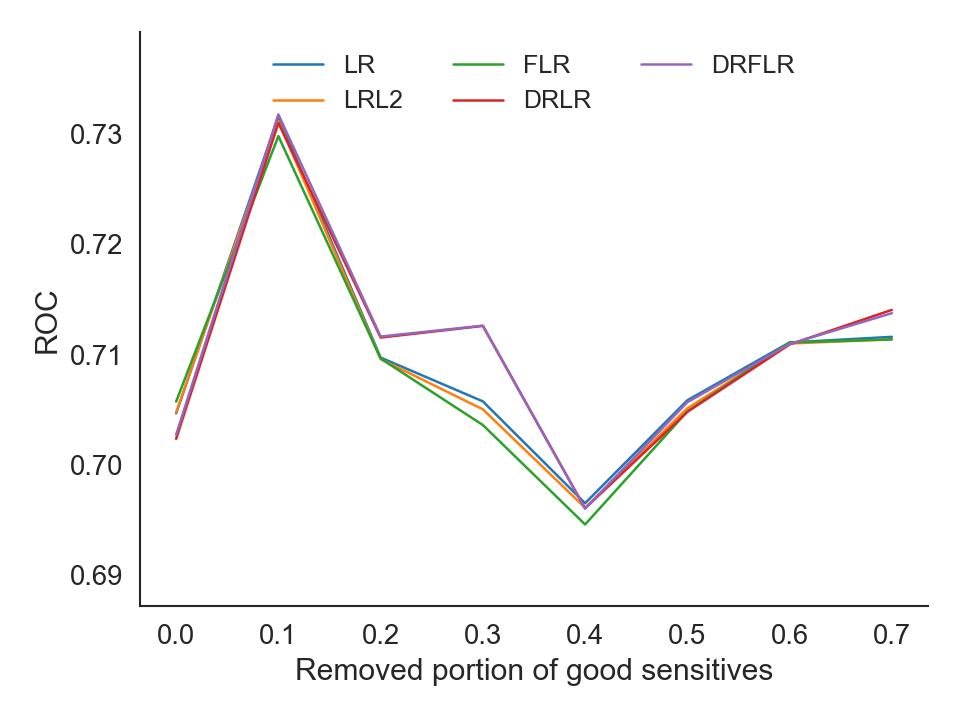} 
    \caption{HC} 
  \end{subfigure} 

  \begin{subfigure}[b]{\multilen}
    \includegraphics[width=0.54\imagewidth]{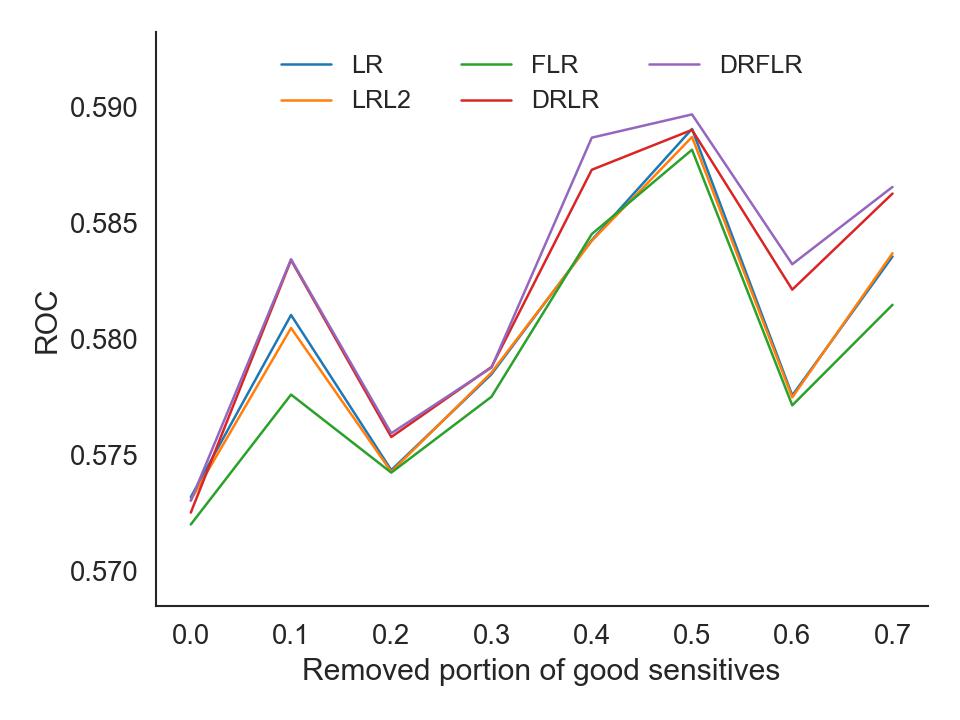} 
    \caption{PAKDD} 
  \end{subfigure} 
  \hfill
  \begin{subfigure}[b]{\multilen}
    \includegraphics[width=0.54\imagewidth]{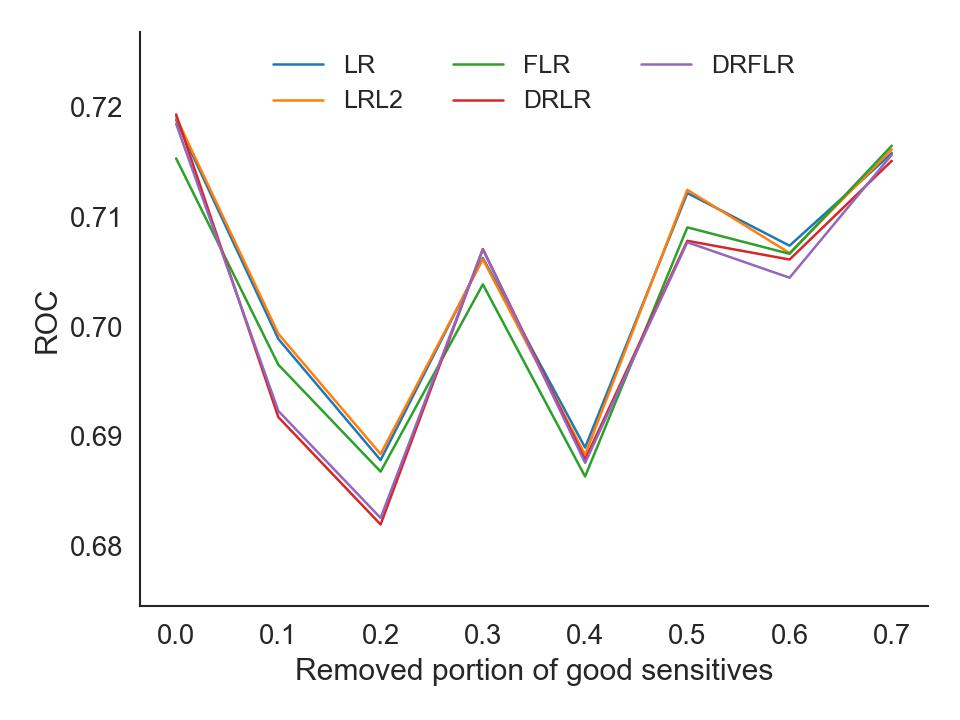} 
    \caption{TC} 
  \end{subfigure} 

  \caption{Evolution of ROC with changes in marginal distribution (GMSC, HC, PAKDD, and TC)}
  \label{appendix:rocprop} 

  \text{\footnotesize$\rho=0.01$, $\kappa_y=\kappa_s=0.4$, $\eta=\frac{\eta_{\max}}{1.5}$}
\end{figure}

\begin{figure}[t] 
  \centering
  \begin{subfigure}[b]{\multilen}
    \includegraphics[width=0.54\imagewidth]{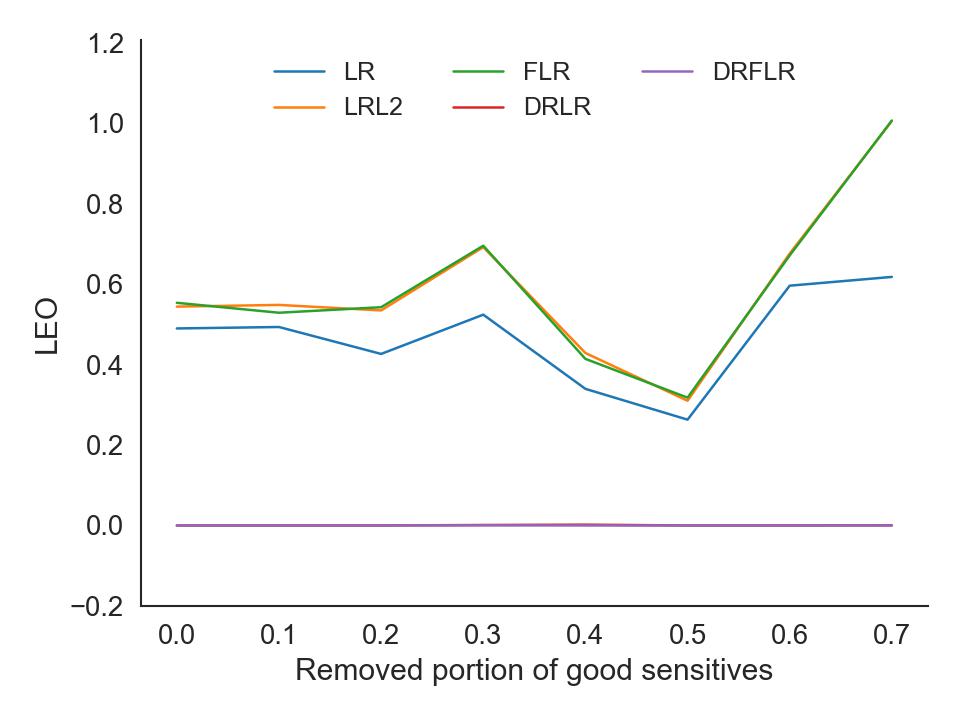}
    \caption{GMSC} 
  \end{subfigure}
  \hfill
  \begin{subfigure}[b]{\multilen}
    \includegraphics[width=0.54\imagewidth]{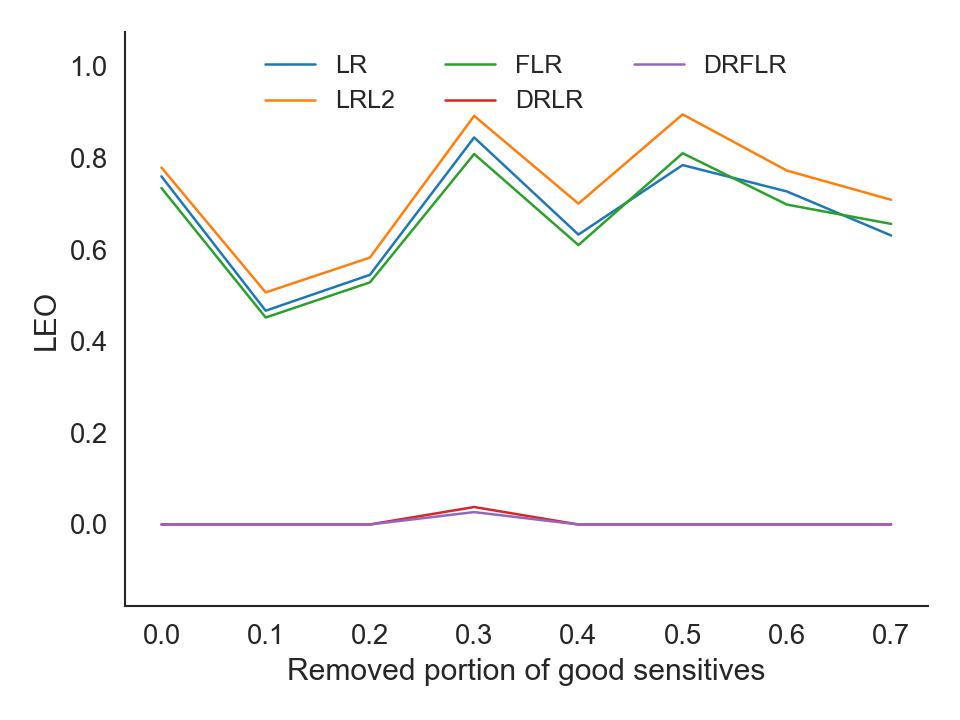} 
    \caption{HC} 
  \end{subfigure} 
 
  \begin{subfigure}[b]{\multilen}
    \includegraphics[width=0.54\imagewidth]{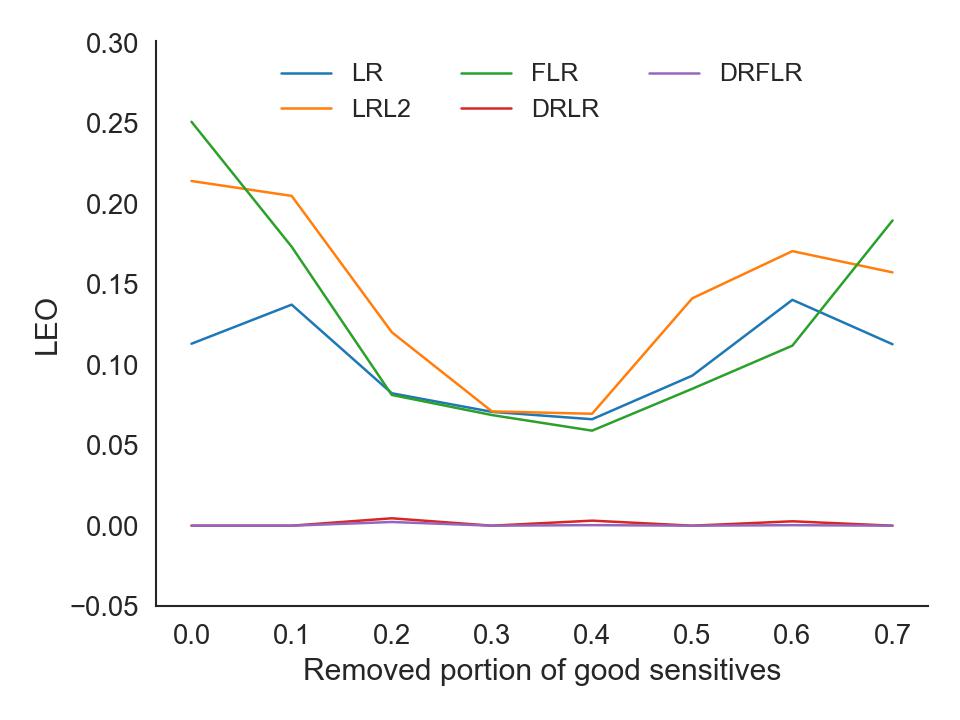} 
    \caption{PAKDD} 
  \end{subfigure} 
  \hfill
  \begin{subfigure}[b]{\multilen}
    \includegraphics[width=0.54\imagewidth]{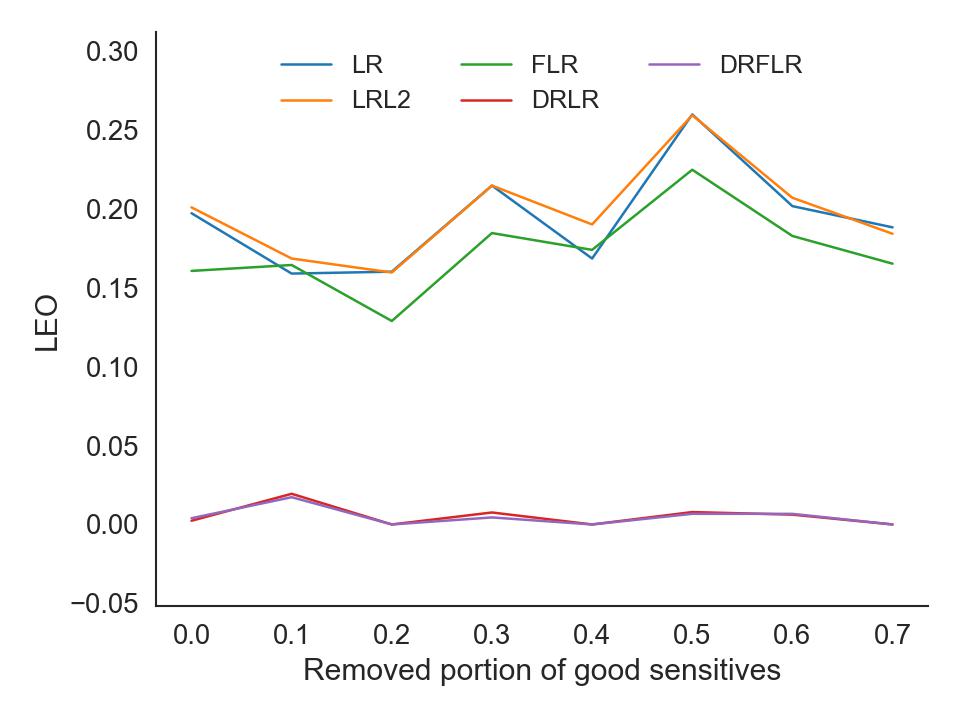} 
    \caption{TC} 
  \end{subfigure} 

  \caption{Evolution of LEO with changes in marginal distribution (GMSC, HC, PAKDD, and TC)}
  \label{appendix:leoprop} 

  \text{\footnotesize$\rho=0.01$, $\kappa_y=\kappa_s=0.4$, $\eta=\frac{\eta_{\max}}{1.5}$}
\end{figure}

While the general trends for GC in Figure \ref{fig:4} appear similar for all algorithms, 
and although there is very little to separate the different methods in terms of ROC, we see both DRFLR and DRLR yielding consistently better LEO than the non-robust methods, while, in some cases, this also applies to SP. A similar observation applies to 
the other datasets (see Figures  \ref{appendix:rocprop} and \ref{appendix:leoprop}). However, it is worth noting that in the case of GMSC, the robust classifiers achieve higher ROC regardless of the level of imbalance (see Figure \ref{appendix:rocprop}a). This further supports the explanation that the performance gains observed there may stem from the feature space heterogeneity. 

\vspace{-2pt}\section{Conclusions and future work}

This paper set out to investigate how to apply DRO methods, either with or without an added fairness penalty, to credit scoring. Having put forward a candidate distributionally robust logistic regression method from the literature, we empirically evaluated how its different components and our proposed hyperparameter tuning strategy impacted predictive performance and fairness, using samples of publicly available real-world credit data. We thus found that distributionally robust methods can indeed provide a substantial improvement in terms of fairness compared to their non-robust equivalents, with little to no loss in performance. These new insights indicate that DRO has the potential to improve fairness in credit scoring, provided that further computational advances are made in efficiently implementing such systems. Indeed, a significant hurdle to their broader adoption by lenders is the high computational cost of the available optimisation algorithms, which exhibit limited scalability with increasing sample sizes and are harder to parallelise efficiently than standard logistic regression. 


Interestingly, our experiments showed a potential connection between robustness and fairness as, even without an explicit fairness penalty being added, DRLR often produced fairer solutions than the non-robust methods. This suggests that increasing robustness and improving generalisation on their own can already have a positive impact on fairness. This connection requires additional research to be fully understood. An interesting direction for future work is to consider the choice of ground metric, for example, by exploring the idea of treating potential proxy features as more distributionally ambiguous than other features. 
%
%
Finally, we argue that there is a need for unified fairness metrics in credit scoring, as most existing metrics either favour low dispersion in predicted PDs or are threshold-dependent, and may therefore provide an overly favourable assessment of fairness under certain conditions and lead to suboptimal model selection. 

\bibliographystyle{elsarticle-harv} 
{\small \bibliography{drfo.bib}}

@article{soyster1973convex,
  title={Convex programming with set-inclusive constraints and applications to inexact linear programming},
  author={Soyster, Allen L},
  journal={Operations Research},
  volume={21},
  number={5},
  pages={1154--1157},
  year={1973},
  publisher={INFORMS}
}

@article{bertsimas2004price,
  title={The price of robustness},
  author={Bertsimas, Dimitris and Sim, Melvyn},
  journal={Operations Research},
  volume={52},
  number={1},
  pages={35--53},
  year={2004},
  publisher={Informs}
}

@article{ben1998robust,
  title={Robust convex optimization},
  author={Ben-Tal, Aharon and Nemirovski, Arkadi},
  journal={Mathematics of Operations Research},
  volume={23},
  number={4},
  pages={769--805},
  year={1998},
  publisher={INFORMS}
}

@techreport{garcia2015creditf,
  title={Credit expansion in emerging markets: propeller of growth?},
  author={Garcia-Escribano, Mercedes and Han, Fei},
  year={2015},
  institution={International Monetary Fund},
  type={IMF Working Paper},
  number={WP/15/212}
}

@techreport{shafieezadeh2020wasserstein,
  title={Wasserstein Distributionally Robust Learning},
  author={Shafieezadeh Abadeh, Soroosh},
  year={2020},
  institution={EPFL}
}

@inproceedings{salehi2019impact,
  title={The Impact of Regularization on High-Dimensional Logistic Regression},
  author={Salehi, Fariborz and Abbasi, Ehsan and Hassibi, Babak},
  booktitle={Advances in Neural Information Processing Systems},
  volume={32},
  pages={11982--11992},
  year={2019}
}

@article{long2023robust,
  title={Robust satisficing},
  author={Long, Daniel Zhuoyu and Sim, Melvyn and Zhou, Minglong},
  journal={Operations Research},
  volume={71},
  number={1},
  pages={61--82},
  year={2023},
  publisher={INFORMS}
}

@inproceedings{kamishima2011fairness,
  title={Fairness-aware learning through regularization approach},
  author={Kamishima, Toshihiro and Akaho, Shotaro and Sakuma, Jun},
  booktitle={2011 IEEE 11th {International Conference on Data Mining }Workshops},
  pages={643--650},
  year={2011},
  organization={IEEE}
}

@article{blanchet2019robust,
  title={Robust {W}asserstein profile inference and applications to machine learning},
  author={Blanchet, Jose and Kang, Yang and Murthy, Karthyek},
  journal={Journal of Applied Probability},
  volume={56},
  number={3},
  pages={830--857},
  year={2019},
  publisher={Cambridge University Press}
}

@article{xu2010robust,
  title={Robust regression and lasso},
  author={Xu, Huan and Caramanis, Constantine and Mannor, Shie},
  journal={IEEE Transactions on Information Theory},
  volume={56},
  number={7},
  pages={3561--3574},
  year={2010},
  publisher={IEEE}
}

@article{taskesen2020distributionally,
  title={A Distributionally Robust Approach to Fair Classification},
  author={Ta{\c{s}}kesen, Bahar and Nguyen, Viet Anh and Kuhn, Daniel and Blanchet, Jose},
  journal={arXiv preprint arXiv:2007.09530},
  year={2020}
}

@article{rahimian2019distributionally,
 title={Frameworks and Results in Distributionally Robust Optimization},
   volume={3},
   ISSN={2777-5860},
   url={http://dx.doi.org/10.5802/ojmo.15},
   DOI={10.5802/ojmo.15},
   journal={Open Journal of Mathematical Optimization},
   publisher={MathDoc/Centre Mersenne},
   author={Rahimian, Hamed and Mehrotra, Sanjay},
   year={2022},
   month=July, pages={1–85} }

@incollection{pessach2020algorithmic,
  title={Algorithmic Fairness},
  author={Pessach, Dana and Shmueli, Erez},
  booktitle={Machine Learning for Data Science Handbook},
  editor={Rokach, Lior and Maimon, Oded and Shmueli, Erez},
  pages={867--886},
  year={2023},
  publisher={Springer},
  doi={10.1007/978-3-031-24628-9_37}
}

@article{martinez2019fairness,
  title={Missing the missing values: The ugly duckling of fairness in machine learning},
  author={Mart{\'i}nez-Plumed, Fernando and Ferri, C{\`e}sar Pallar{\'e}s and Nieves, David and Hern{\'a}ndez-Orallo, Jos{\'e}},
  journal={International Journal of Intelligent Systems},
  volume={36},
  number={7},
  pages={3217--3258},
  year={2021},
  publisher={Wiley Online Library}
}

@article{chouldechova2018frontiers,
  title={The Frontiers of Fairness in Machine Learning}, 
  author={Alexandra Chouldechova and Aaron Roth},
  year={2018},
  journal={arXiv preprint arXiv:1810.08810},
  primaryClass={cs.LG},
  url={https://arxiv.org/abs/1810.08810}
}

@article{selbst2017disparate,
  title   = {{Disparate Impact in Big Data Policing}},
  author  = {Selbst, Andrew D.},
  journal = {Georgia Law Review},
  volume  = {52},
  number  = {1},
  pages   = {109--195},
  year    = {2017},
  url     = {https://digitalcommons.law.uga.edu/glr/vol52/iss1/6/}
}

@inproceedings{dwork2012fairness,
  title={Fairness through awareness},
  author={Dwork, Cynthia and Hardt, Moritz and Pitassi, Toniann and Reingold, Omer and Zemel, Richard},
  booktitle={Proceedings of the 3rd {I}nnovations in {T}heoretical {C}omputer {S}cience {C}onference},
  pages={214--226},
  year={2012}
}

@article{chouldechova2017fair,
  title={Fair prediction with disparate impact: A study of bias in recidivism prediction instruments},
  author={Chouldechova, Alexandra},
  journal={Big Data},
  volume={5},
  number={2},
  pages={153--163},
  year={2017},
  publisher={Mary Ann Liebert, Inc. 140 Huguenot Street, 3rd Floor New Rochelle, NY 10801 USA}
}

@article{andreeva2019law,
  title={The law of equal opportunities or unintended consequences?: The effect of unisex risk assessment in consumer credit},
  author={Andreeva, Galina and Matuszyk, Anna},
  journal={Journal of the Royal Statistical Society: Series A (Statistics in Society)},
  volume={182},
  number={4},
  pages={1287--1311},
  year={2019},
  publisher={Wiley Online Library}
}

@article{kozodoi2022fairness,
  title={Fairness in credit scoring: Assessment, implementation and profit implications},
  author={Kozodoi, Nikita and Jacob, Johannes and Lessmann, Stefan},
  journal={European Journal of Operational Research},
  volume={297},
  number={3},
  pages={1083--1094},
  year={2022},
  publisher={Elsevier}
}

@inproceedings{kamiran2009classifying,
  title={Classifying without discriminating},
  author={Kamiran, Faisal and Calders, Toon},
  booktitle={2009 2nd {International Conference on Computer, Control and Communication}},
  pages={1--6},
  year={2009},
  organization={IEEE}
}

@article{ovadia2019can,
  title={Can you trust your model's uncertainty? Evaluating predictive uncertainty under dataset shift},
  author={Ovadia, Yaniv and Fertig, Emily and Ren, Jie and Nado, Zachary and Sculley, David and Nowozin, Sebastian and Dillon, Joshua and Lakshminarayanan, Balaji and Snoek, Jasper},
  journal={Advances in Neural Information Processing Systems},
  volume={32},
  year={2019}
}

@inproceedings{subbaswamy2021evaluating,
  title={Evaluating model robustness and stability to dataset shift},
  author={Subbaswamy, Adarsh and Adams, Roy and Saria, Suchi},
  booktitle={International Conference on Artificial Intelligence and Statistics},
  pages={2611--2619},
  year={2021},
  organization={PMLR}
}

@inproceedings{krempl2011classification,
  title={Classification in presence of drift and latency},
  author={Krempl, Georg and Hofer, Vera},
  booktitle={2011 IEEE 11th International Conference on Data Mining Workshops},
  pages={596--603},
  year={2011},
  organization={IEEE}
}

@misc{falque-pierrotin_2017,
  author={{Article 29 Data Protection Working Party (WP29)}},
  title={Guidelines on Data Protection Officers ({`DPOs'}) (WP243 rev.01)},
  url={https://ec.europa.eu/newsroom/article29/items/612048/en},
  publisher={{European Commission}},
  year={2017},
  
}

@techreport{muñoz_smith_patil_2016,
  title={Big Data: A Report on Algorithmic Systems, Opportunity, and Civil Rights},
  author={{Executive Office of the President}},
  institution={The White House},
  year={2016},
  url={https://obamawhitehouse.archives.gov/sites/default/files/microsites/ostp/2016_0504_data_discrimination.pdf}
}

@book{barocas2017fairness, title={Fairness and machine learning: Limitations and opportunities}, author={Barocas, Solon and Hardt, Moritz and Narayanan, Arvind}, year={2023}, publisher={MIT Press} }

@inproceedings{calders2009building,
  title={Building classifiers with independency constraints},
  author={Calders, Toon and Kamiran, Faisal and Pechenizkiy, Mykola},
  booktitle={2009 IEEE International Conference on Data Mining Workshops},
  pages={13--18},
  year={2009},
  organization={IEEE}
}

@inproceedings{kamiran2012decision,
  title={Decision theory for discrimination-aware classification},
  author={Kamiran, Faisal and Karim, Asim and Zhang, Xiangliang},
  booktitle={2012 IEEE 12th {International Conference on Data Mining}},
  pages={924--929},
  year={2012},
  organization={IEEE}
}

@misc{Dua:2019 ,
author       = {Dua, Dheeru and Graff, Casey},
  year         = {2017},
  title        = {{UCI} Machine Learning Repository},
  url          = {http://archive.ics.uci.edu/ml},
  howpublished = {University of California, Irvine, School of Information and Computer Sciences}}

@misc{i-cheng_2009,
  author       = {I-Cheng Yeh},
  title        = {Default of Credit Card Clients Data Set},
  year         = {2016},
  url          = {https://archive.ics.uci.edu/dataset/350/default+of+credit+card+clients},
  howpublished = {UCI Machine Learning Repository}
}

@misc{hofmann1994, title={ Statlog {(German Credit Data)} Data Set}, url={https://archive.ics.uci.edu/dataset/144/statlog+german+credit+data}, author={Hofmann, Hans}, year={1994}}

@misc{GiveMeSomeCredit,
 author       = {{Credit Fusion} and Will Cukierski},
  title        = {Give {Me} {Some} {Credit}},
  year         = {2011},
  howpublished = {Kaggle Competition},
  url          = {https://kaggle.com/competitions/GiveMeSomeCredit},
}

@article{tian2022comprehensive,
  author = {Tian, Yingjie and Zhang, Yuqi},
  title = {A comprehensive survey on regularization strategies in machine learning},
  journal = {Information Fusion},
  volume = {80},
  pages = {146--166},
  year = {2022}
}

@inproceedings{wang2020robust,
  title={Robust Optimization for Fairness with Noisy Protected Groups},
  author={Wang, Serena and Guo, Wenshuo and Narasimhan, Harikrishna and Cotter, Andrew and Gupta, Maya and Jordan, Michael I.},
  booktitle={Advances in Neural Information Processing Systems},
  volume={33},
  pages={5190--5203},
  year={2020}
}

@article{cuturi2014ground,
  title={Ground metric learning},
  author={Cuturi, Marco and Avis, David},
  journal={The Journal of Machine Learning Research},
  volume={15},
  number={1},
  pages={533--564},
  year={2014},
  publisher={JMLR. org}
}

@article{shafieezadeh2015distributionally,
  title={Distributionally robust logistic regression},
  author={Shafieezadeh Abadeh, Soroosh and Mohajerin Esfahani, Peyman M and Kuhn, Daniel},
  journal={Advances in Neural Information Processing Systems},
  volume={28},
  year={2015}
}

@article{li2019first,
  title={A First-Order Algorithmic Framework for Distributionally Robust Logistic Regression},
  author={Li, Jiajin and Huang, Sen and So, Anthony Man-Cho},
  journal={Advances in Neural Information Processing Systems},
  volume={32},
  year={2019}
}

@incollection{vzliobaite2016overview,
  title={An Overview of Concept Drift Applications},
  author={{\v{Z}}liobait{\.{e}}, Indr{\.{e}} and Pechenizkiy, Mykola and Gama, Jo{\~a}o},
  booktitle={Big Data Analysis: New Algorithms for a New Society},
  editor={Japkowicz, Nathalie and Stefanowski, Jerzy},
  series={Studies in Big Data},
  volume={16},
  pages={91--114},
  year={2016},
  publisher={Springer},
  address={Cham}
}

@book{thomas_edelman_croock_2002, title={Credit Scoring and Its Applications},
  author={Thomas, Lyn C. and Edelman, David B. and Crook, Jonathan N.},
  publisher={Society for Industrial and Applied Mathematics},
  address={Philadelphia, PA},
  year={2002}
}

@book{ben2009robust,
  title={Robust optimization},
  author={Ben-Tal, Aharon and El Ghaoui, Laurent and Nemirovski, Arkadi},
 
  year={2009},
  publisher={Princeton University Press}
}

@book{murphy2012machine,
  title={Machine learning: a probabilistic perspective},
  author={Murphy, Kevin P},
  year={2012},
  publisher={MIT press}
}

@article{moscato2021benchmark,
  title={A benchmark of machine learning approaches for credit score prediction},
  author={Moscato, Vincenzo and Picariello, Antonio and Sperli, Giancarlo},
  journal={Expert Systems with Applications},
  volume={165},
  pages={113986},
  year={2021},
  publisher={Elsevier}
}

@article{lessmann2015benchmarking,
  title={Benchmarking state-of-the-art classification algorithms for credit scoring: An update of research},
  author={Lessmann, Stefan and Baesens, Bart and Seow, Hsin-Vonn and Thomas, Lyn C},
  journal={European Journal of Operational Research},
  volume={247},
  number={1},
  pages={124--136},
  year={2015},
  publisher={Elsevier}
}

@inproceedings{ying2019overview,
  title={An overview of overfitting and its solutions},
  author={Ying, Xue},
  booktitle={Journal of Physics: Conference series},
  volume={1168},
  pages={022022},
  year={2019},
  organization={IOP Publishing}
}

@article{kantorovich1960mathematical,
  title={Mathematical methods of organizing and planning production},
  author={Kantorovich, Leonid V},
  journal={Management science},
  volume={6},
  number={4},
  pages={366--422},
  year={1960},
  publisher={INFORMS}
}

@article{gao2022wasserstein,
  title={Wasserstein distributionally robust optimization and variation regularization},
  author={Gao, Rui and Chen, Xi and Kleywegt, Anton J},
  journal={Operations Research},
  volume={72},
  number={3},
  pages={1177--1191},
  year={2024},
  publisher={INFORMS}
}

@inproceedings{NIPS2016_9d268236,
  author    = {Hardt, Moritz and Price, Eric and Srebro, Nati},
  booktitle = {Advances in Neural Information Processing Systems},
  editor    = {D. Lee and M. Sugiyama and U. Luxburg and I. Guyon and R. Garnett},
  pages     = {3323--3331},
  publisher = {Curran Associates, Inc.},
  title     = {Equality of Opportunity in Supervised Learning},
  url       = {https://proceedings.neurips.cc/paper_files/paper/2016/file/9d2682367c3935defcb1f9e247a97c0d-Paper.pdf},
  volume    = {29},
  year      = {2016}
}

@inproceedings{ealigam, 
  title={The 14th {P}acific-{A}sia {C}onference on {K}nowledge {D}iscovery and {D}ata mining}, 
  url={https://pakdd.org/archive/pakdd2010/PAKDDCompetition.html}, 
  journal={PAKDD 2010}, 
  author={{PAKDD 2010 Data Mining Competition / Organising Committee}},
  year = {2010}
}

@misc{homecr,
  author = {{Kaggle / Home Credit Group}},
  title = {Home Credit Default Risk},
  year = {2018},
  howpublished = {\url{https://kaggle.com/competitions/home-credit-default-risk}},
  note = {Kaggle Competition}
}

@article{youden1950index,
  title={Index for rating diagnostic tests},
  author={Youden, William J},
  journal={Cancer},
  volume={3},
  number={1},
  pages={32--35},
  year={1950},
  publisher={Wiley Online Library},
  doi={10.1002/1097-0142(1950)3:1<32::aid-cncr2820030106>3.0.co;2-3}
}

@article{tibshirani1996regression,
  title={Regression shrinkage and selection via the lasso},
  author={Tibshirani, Robert},
  journal={Journal of the Royal Statistical Society Series B: Statistical Methodology},
  volume={58},
  number={1},
  pages={267--288},
  year={1996},
  publisher={Oxford University Press}
}

@inproceedings{petersen_post-processing_2021,
	title={Post-processing for Individual Fairness},
  author={Petersen, Felix and Mukherjee, Debarghya and Sun, Yuekai and Yurochkin, Mikhail},
  booktitle={Advances in Neural Information Processing Systems},
  volume={34},
  pages={25964--25976},
  year={2021}
}

@article{wan_-processing_2023,
  title={In-Processing Modeling Techniques for Machine Learning Fairness: A Survey},
  author={Wan, Mingyang and Zha, Daochen and Liu, Ninghao and Zou, Na},
  journal={ACM Transactions on Knowledge Discovery from Data},
  volume={17},
  number={3},
  pages={1--27},
  year={2023},
  doi={10.1145/3551390}
}

@inproceedings{zhang_review_2023,
	location = {Cham},
	title = {A Review on Pre-processing Methods for Fairness in Machine Learning},
	isbn = {978-3-031-20738-9},
	pages = {1185--1191},
    year={2023},
	booktitle = {Advances in Natural Computation, Fuzzy Systems and Knowledge Discovery},
	publisher = {Springer International Publishing},
	author = {Zhang, Zhe and Wang, Shenhang and Meng, Gong},
	editor = {Xiong, Ning and Li, Maozhen and Li, Kenli and Xiao, Zheng and Liao, Longlong and Wang, Lipo},
	date = {2023},
}

@article{smirnov1948table,
 title={Table for estimating the goodness of fit of empirical distributions},
  author={Smirnov, Nickolay},
  journal={The Annals of Mathematical Statistics},
  volume={19},
  pages={279--281},
  year={1948},
  publisher={Institute of Mathematical Statistics}
}

@article{kolmogorov1933,
  author    = {Kolmogorov, Andrey N.},
  title     = {Sulla determinazione empirica di una legge di distribuzione},
  journal   = {Giornale dell'Istituto Italiano degli Attuari},
  year      = {1933},
  volume    = {4},
  pages     = {83--91},
  language  = {italian}
}

@ARTICLE{2020SciPy-NMeth,
  author  = {Virtanen, Pauli and Gommers, Ralf and Oliphant, Travis E. and
            Haberland, Matt and Reddy, Tyler and Cournapeau, David and
            Burovski, Evgeni and Peterson, Pearu and Weckesser, Warren and
            Bright, Jonathan and {van der Walt}, St{\'e}fan J. and
            Brett, Matthew and Wilson, Joshua and Millman, K. Jarrod and
            Mayorov, Nikolay and Nelson, Andrew R. J. and Jones, Eric and
            Kern, Robert and Larson, Eric and Carey, C J and
            Polat, {\.I}lhan and Feng, Yu and Moore, Eric W. and
            {VanderPlas}, Jake and Laxalde, Denis and Perktold, Josef and
            Cimrman, Robert and Henriksen, Ian and Quintero, E. A. and
            Harris, Charles R. and Archibald, Anne M. and
            Ribeiro, Ant{\^o}nio H. and Pedregosa, Fabian and
            {van Mulbregt}, Paul and {SciPy 1.0 Contributors}},
  title   = {{{SciPy} 1.0: Fundamental Algorithms for Scientific
            Computing in Python}},
  journal = {Nature Methods},
  year    = {2020},
  volume  = {17},
  pages   = {261--272},
  adsurl  = {https://rdcu.be/b08Wh},
  doi     = {10.1038/s41592-019-0686-2},
}

@article{hurlin2026fairness,
  title={The fairness of credit scoring models},
  author={Hurlin, Christophe and P{\'e}rignon, Christophe and Saurin, S{\'e}bastien},
  journal={Management Science},
  volume={72},
  number={1},
  pages={406--425},
  year={2026},
  publisher={INFORMS}
}

@article{wang2024wasserstein,
  title={Wasserstein robust classification with fairness constraints},
  author={Wang, Yijie and Nguyen, Viet Anh and Hanasusanto, Grani A},
  journal={Manufacturing \& Service Operations Management},
  volume={26},
  number={4},
  pages={1567--1585},
  year={2024},
  publisher={INFORMS}
}

@article{slowik2021algorithmic,
    title={Algorithmic Bias and Data Bias: Understanding the Relation between Distributionally Robust Optimization and Data Curation},
  author={S{\l}owik, Agnieszka and Bottou, L{\'e}on},
  journal={arXiv preprint arXiv:2106.09467},
  year={2021}
}

@article{kim2026fair,
  title={Fair models in credit: Intersectional discrimination and the amplification of inequity},
  author={Kim, Savina and Lessmann, Stefan and Andreeva, Galina and Rovatsos, Michael},
  journal={Annals of Operations Research},
  volume={361},
  number={1},
  pages={465--502},
  year={2026},
  publisher={Springer}
}

@inproceedings{sargeant2025formalising,
  title={Formalising anti-discrimination law in automated decision systems},
  author={Sargeant, Holli and Magnusson, M{\aa}ns},
  booktitle={Proceedings of the 2025 ACM Conference on Fairness, Accountability, and Transparency},
  pages={181--194},
  year={2025}
}

@INPROCEEDINGS{11401835,
  author={Nisha, M. Reema and Amirtha Varshini, R. and Aswath, S. and Basavaraj, K. and Dharshini, G. L. and Aruna, N.},
  booktitle={2025 IEEE First International Conference on Innovations in Engineering and Next-Generation Technologies for Sustainability (ICINVENTS)}, 
  title={Trustworthy Credit Scoring: Fairness and Explainability in Financial {AI}}, 
  year={2025},
  volume={1},
  number={},
  pages={1--6},
  doi={10.1109/ICINVENTS64613.2025.11401835}
}

\appendix
\makeatletter
\def\thesection{\appendixname} 
\makeatother

\section*{Appendix}
To implement the DRO problem with the added fairness penalty, we use the tractable reformulation of problem \eqref{DRFLR} by \citet{taskesen2020distributionally}:  
\begin{align}
\text{min:} \quad & t \\
\text{s.t.:}\footnotesize \quad & \mathbf{w} \in \mathbb{R}^m, \psi_s \in \mathbb{R}_+, d \in \mathbb{R}^n, \mu \in \mathbb{R}^{2 \times 2}\normalsize \\
\text{} \quad & \rho \psi_s + \frac{1}{n} \sum_{i=1}^{n} d_{s'i} + \sum_{s \in \{0,1\}} \sum_{y \in \{0,1\}} \hat{p}_{sy} \mu_{s'sy} \leq t \quad \forall s,s' \in \{0,1\}, \,s'=1-s\\
&\begin{aligned}
& \|\mathbf{w}\|_*(1+\eta/\hat{p}_{01}) \leq \psi_0; & \quad \|\mathbf{w}\|_*(1+\eta/\hat{p}_{11}) \leq \psi_1 \\
\end{aligned}\label{eq16_2}
\\
& \begin{aligned}
& \left. \begin{aligned}
& -\ln(1-h(\mathbf{x}_i;\mathbf{w}))-\psi_s\kappa_s|s-s_i| -\psi_s\kappa_y|y_i|-\mu_{s0} \leq d_{s i}, \\
& -\ln(1-h(\mathbf{x}_i;\mathbf{w}))-\psi_s\kappa_s|s'-s_i| -\psi_s\kappa_y|y_i|-\mu_{s'0} \leq d_{s i}, \\
& -(1-\dfrac{\eta}{\hat{p}_{s1}})\ln(h(\mathbf{x}_i;\mathbf{w}))-\psi_s\kappa_s|s-s_i| -\psi_s\kappa_y|1-y_i|-\mu_{s1} \leq d_{s i},  \\
& -(1-\dfrac{\eta}{\hat{p}_{s'1}})\ln(h(\mathbf{x}_i;\mathbf{w}))-\psi_s\kappa_s|s'-s_i| -\psi_s\kappa_y|1-y_i|-\mu_{s'1} \leq d_{s i}
\end{aligned} \right) \scriptsize\begin{aligned}{\forall i \leq n}\\{ \forall s \in \{0,1\}}\\ s'=1-s\end{aligned}
\end{aligned}
\end{align}
\normalsize
where $\| \cdot \|_*$ represents the norm dual to $\|\cdot\|$, \(t\) is an epigraphical variable, and \(d\), \(\psi\), and \(\mu\) are dual variables.
\end{document}